# Faster Principal Component Regression and Stable Matrix Chebyshev Approximation

Zeyuan Allen-Zhu
zeyuan@csail.mit.edu
Princeton University / IAS

Yuanzhi Li
yuanzhil@cs.princeton.edu
Princeton University

**Abstract**

We solve principal component regression (PCR), up to a multiplicative accuracy $1+\gamma$, by reducing the problem to $\widetilde{O}(\gamma^{-1})$ black-box calls of ridge regression. Therefore, our algorithm does not require any explicit construction of the top principal components, and is suitable for large-scale PCR instances. In contrast, previous result requires $\widetilde{O}(\gamma^{-2})$ such black-box calls.

We obtain this result by developing a general stable recurrence formula for matrix Chebyshev polynomials, and a degree-optimal polynomial approximation to the matrix sign function. Our techniques may be of independent interests, especially when designing iterative methods.

## 1 Introduction

In machine learning and statistics, it is often desirable to represent a large-scale dataset in a more tractable, lower-dimensional form, without losing too much information. One of the most robust ways to achieve this goal is through *principal component projection (PCP)*:

PCP: project vectors onto the span of the top principal components of the a matrix.

It is well-known that PCP decreases noise and increases efficiency in downstream tasks. One of the main applications is *principal component regression (PCR)*:

PCR: linear regression but restricted to the subspace of top principal components.

Classical algorithms for PCP or PCR rely on a principal component analysis (PCA) solver to recover the top principal components first; with these components available, the tasks of PCP and PCR become trivial because the projection matrix can be constructed explicitly.

Unfortunately, PCA solvers demand a running time that at least linearly scales with the number of top principal components chosen for the projection. For instance, to project a vector onto the top 1000 principal components of a high-dimensional dataset, even the most efficient Krylov-based [18] or Lanczos-based [4] methods require a running time that is proportional to $1000 \times 40 = 4 \times 10^4$ times the input matrix sparsity, if the Krylov or Lanczos method is executed for 40 iterations. This is usually computationally intractable.

## 1.1 Approximating PCP Without PCA

In this paper, we propose the following notion of PCP approximation. Given a data matrix $\mathbf{A} \in \mathbb{R}^{d' \times d}$ (with singular values no greater than 1) and a threshold $\lambda > 0$, we say that an algorithm solves $(\gamma, \varepsilon)$-approximate PCP if —informally speaking and up to a multiplicative $1 \pm \varepsilon$ error— it projects (see Def. 3.1 for a formal definition)

1. any eigenvector $\nu$ of $\mathbf{A}^\top \mathbf{A}$ with value in $\big[\lambda(1+\gamma), 1\big]$ to $\nu$,
2. any eigenvector $\nu$ of $\mathbf{A}^\top \mathbf{A}$ with value in $\big[0, \lambda(1-\gamma)\big]$ to $\vec{0}$,
3. any eigenvector $\nu$ of $\mathbf{A}^\top \mathbf{A}$ with value in $\big[\lambda(1-\gamma), \lambda(1+\gamma)\big]$ to "anywhere between $\vec{0}$ and $\nu$."

Such a definition also extends to $(\gamma, \varepsilon)$-approximate PCR (see Def. 3.2).

It was first noticed by Frostig *et al.* [13] that approximate PCP and PCR be solved with a running time independent of the number of principal components above threshold $\lambda$. More specifically, they reduced $(\gamma, \varepsilon)$-approximate PCP and PCR to

$$\boxed{O\big(\gamma^{-2} \log(1/\varepsilon)\big) \text{ black-box calls of any ridge regression subroutine}}$$

where each call computes $(\mathbf{A}^\top \mathbf{A} + \lambda \mathbf{I})^{-1} u$ for some vector $u$.[1] Our main focus of this paper is to *quadratically* improve this performance and reduce PCP and PCR to

$$\boxed{O\big(\gamma^{-1} \log(1/\gamma\varepsilon)\big) \text{ black-box calls of any ridge regression subroutine}}$$

where each call again computes $(\mathbf{A}^\top \mathbf{A} + \lambda \mathbf{I})^{-1} u$.

*Remark* 1.1. Frostig *et al.* only showed their algorithm satisfies the properties 1 and 2 of $(\gamma, \varepsilon)$-approximation (but not the property 3), and thus their proof was only for matrix $\mathbf{A}$ with no singular value in the range $[\sqrt{\lambda(1-\gamma)}, \sqrt{\lambda(1+\gamma)}]$. This is known as the *eigengap assumption*, which is rarely satisfied in practice [18]. In this paper, we prove our result both with and without such eigengap assumption. Since our techniques also imply the algorithm of Frostig *et al.* satisfies property 3, throughout the paper, we say Frostig *et al.* solve $(\gamma, \varepsilon)$-approximate PCP and PCR.

## 1.2 From PCP to Polynomial Approximation

The main technique of Frostig *et al.* is to construct a *polynomial* to approximate the sign function $\operatorname{sgn}(x) \colon [-1, 1] \to \{\pm 1\}$:

$$\operatorname{sgn}(x) \stackrel{\text{def}}{=} \begin{cases} +1, & x \geq 0; \\ -1, & x < 0. \end{cases}$$

In particular, given any polynomial $g(x)$ satisfying

$$\big|g(x) - \operatorname{sgn}(x)\big| \leq \varepsilon \quad \forall x \in [-1, -\gamma] \cup [\gamma, 1] \ , \text{ and} \tag{1.1}$$

$$\big|g(x)\big| \leq 1 \quad \forall x \in [-\gamma, \gamma] \ , \tag{1.2}$$

the problem of $(\gamma, \varepsilon)$-approximate PCP can be reduced to computing the matrix polynomial $g(\mathbf{S})$ for $\mathbf{S} \stackrel{\text{def}}{=} (\mathbf{A}^\top \mathbf{A} + \lambda \mathbf{I})^{-1} (\mathbf{A}^\top \mathbf{A} - \lambda \mathbf{I})$ (cf. Fact 7.1). In other words,

- to project any vector $\chi \in \mathbb{R}^d$ to top principal components, we can compute $g(\mathbf{S})\chi$ instead; and

[1] Ridge regression is often considered as an easy-to-solve machine learning problem: using for instance SVRG [17], one can usually solve ridge regression to an $10^{-8}$ accuracy with at most 40 passes of the data.

- to compute $g(\mathbf{S})\chi$, we can reduce it to ridge regression for each evaluation of $\mathbf{S}u$ for some vector $u$.

*Remark* 1.2. Since the transformation from $\mathbf{A}^\top\mathbf{A}$ to $\mathbf{S}$ is not linear, the final approximation to the PCP is a rational function (as opposed to a polynomial) over $\mathbf{A}^\top\mathbf{A}$. We restrict to polynomial choices of $g(\cdot)$ because in this way, the final rational function has all the denominators being $\mathbf{A}^\top\mathbf{A} + \lambda\mathbf{I}$, thus reduces to ridge regressions.

*Remark* 1.3. The transformation from $\mathbf{A}^\top\mathbf{A}$ to $\mathbf{S}$ ensures that all the eigenvalues of $\mathbf{A}^\top\mathbf{A}$ in the range $(1 \pm \gamma)\lambda$ roughly map to the eigenvalues of $\mathbf{S}$ in the range $[-\gamma, \gamma]$.

**Main Challenges.** There are two main challenges regarding the design of polynomial $g(x)$.

- Efficiency. We wish to minimize the degree $n = \deg(g(x))$ because the computation of $g(\mathbf{S})\chi$ usually requires $n$ calls of ridge regression.
- Stability. We wish $g(x)$ to be stable; that is, $g(\mathbf{S})\chi$ must be given by a recursive formula where if we make $\varepsilon'$ error in each recursion (due to error incurred from ridge regression), the final error of $g(\mathbf{S})\chi$ must be at most $\varepsilon' \times \mathsf{poly}(d)$.

*Remark* 1.4. Efficient routines such as SVRG [17] solve ridge regression and thus compute $\mathbf{S}u$ for any $u \in \mathbb{R}^d$, with running times only logarithmically in $1/\varepsilon'$. Therefore, by setting $\varepsilon' = \varepsilon/\mathsf{poly}(d)$, one can blow up the running time by a small factor $O(\log(d))$ in order to obtain an $\varepsilon$-accurate solution for $g(\mathbf{S})\chi$.

The polynomial $g(x)$ constructed by Frostig *et al.* comes from truncated Taylor expansion. It has degree $O\big(\gamma^{-2}\log(1/\varepsilon)\big)$ and is stable. This $\gamma^{-2}$ dependency limits the practical performance of their proposed PCP and PCR algorithms, especially in a high accuracy regime. At the same time,

- the optimal degree for a polynomial to satisfy even only (1.1) is $\Theta\big(\gamma^{-1}\log(1/\varepsilon)\big)$ [9, 10].

Frostig *et al.* were unable to find a stable polynomial matching this optimal degree and left it as open question.[2]

## 1.3 Our Results and Main Ideas

We provide an efficient and stable polynomial approximation to the matrix sign function that has a near-optimal degree $O(\gamma^{-1}\log(1/\gamma\varepsilon))$. At a high level, we construct a polynomial $q(x)$ that approximately equals $\big(\frac{1+\kappa-x}{2}\big)^{-1/2}$ for some $\kappa = \Theta(\gamma^2)$; then we set $g(x) \stackrel{\text{def}}{=} x \cdot q(1+\kappa-2x^2)$ which approximates $\mathrm{sgn}(x)$.

To construct $q(x)$, we first note that $\big(\frac{1+\kappa-x}{2}\big)^{-1/2}$ has no singular point on $[-1, 1]$ so we can apply Chebyshev approximation theory to obtain some $q(x)$ of degree $O(\gamma^{-1}\log(1/\gamma\varepsilon))$ satisfying

$$\left|q(x) - \Big(\frac{1+\kappa-x}{2}\Big)^{-1/2}\right| \leq \varepsilon \text{ for every } x \in [-1, 1] \quad .$$

This can be shown to imply $\big|g(x) - \mathrm{sgn}(x)\big| \leq \varepsilon$ for every $x \in [-1, -\gamma] \cup [\gamma, 1]$, so (1.1) is satisfied.

In order to prove (1.2) (i.e., $\big|g(x)\big| \leq 1$ for every $x \in [-\gamma, \gamma]$) , we prove a separate lemma:[3]

$$q(x) \leq \Big(\frac{1+\kappa-x}{2}\Big)^{-1/2} \text{ for every } x \in [1, 1+\kappa] \quad .$$

[2]Using degree reduction, Frostig *et al.* found an explicit polynomial $g(x)$ of degree $O\big(\gamma^{-1}\log(1/\gamma\varepsilon)\big)$ satisfying (1.1). However, that polynomial is unstable because it is constructed monomial by monomial and has exponentially large coefficients in front of each monomial. Furthermore, it is not clear if their polynomial satisfies the (1.2).

[3]We proved a general lemma which holds for any function whose all orders of derivatives are non-negative at $x = 0$.

Note that this does not follow from standard Chebyshev theory because Chebyshev approximation guarantees are only with respect to $x \in [-1, 1]$ and do not extend to singular point $x = 1 + \kappa$.

This proves the "EFFICIENCY" part of the main challenges discussed earlier. As for the "STABILITY" part, we prove a general theorem regarding any weighted sum of Chebyshev polynomials applied to matrices. We provide a backward recurrence algorithm and show that it is stable under noisy computations. This may be of independent interest.

For interested readers, we compare our polynomial $q(x)$ with that of Frostig *et al.* in Figure 1.

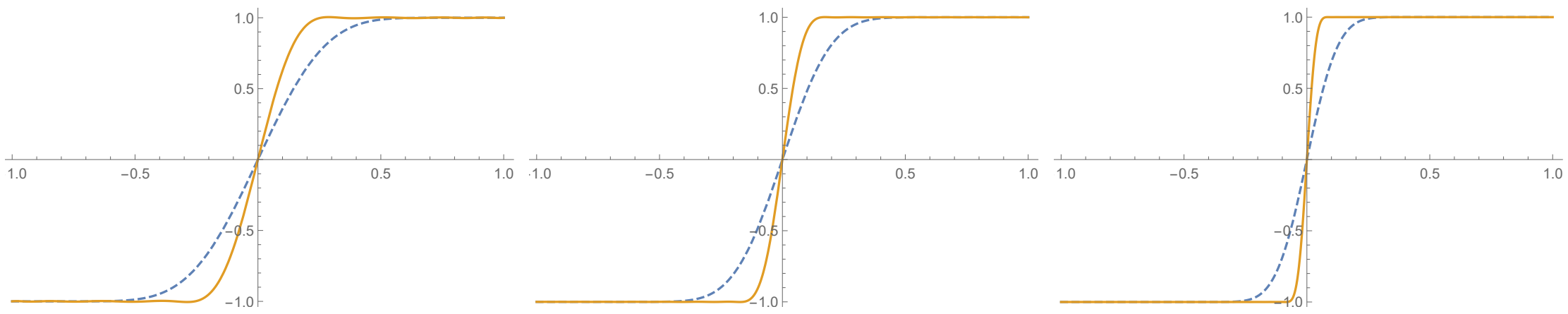

(a) degree 21 (b) degree 41 (c) degree 101

Figure 1: Comparing our polynomial $g(x)$ (orange solid curve) with that of Frostig *et al.* (blue dashed curve).

## 1.4 Related Work

There are a few attempts to reduce the cost of PCA when solving PCR, by for instance approximating the matrix $\mathbf{AP}_\lambda$ where $\mathbf{P}_\lambda$ is the PCP projection matrix [6, 7]. However, they cost a running time that linearly scales with the number of principal components above $\lambda$.

A significant number of papers have focused on the low-rank case of PCA [2, 4, 18] and its online variant [3]. Unfortunately, all of these methods require a running time that scales at least linearly with respect to the number of top principal components.

More related to this paper is work on *matrix sign function*, which plays an important role in control theory and quantum chromodynamics. Several results have addressed Krylov methods for applying the sign function in the so-called Krylov subspace, without explicitly constructing any approximate polynomial [21, 24]. However, Krylov methods are not $(\gamma, \varepsilon)$-approximate PCP solvers, and there is no supporting stability theory behind them.[4] Other iterative methods have also been proposed, see Section 5 of textbook [16]. For instance, Schur's method is a slow one and also requires the matrix to be explicitly given. The Newton's iteration and its numerous variants (e.g. [19]) provide rational approximations to the matrix sign function as opposed to polynomial approximations. Our result and Frostig *et al.* [13] differ from these cited works, because we have only accessed an *approximate ridge regression* oracle, so ensuring a *polynomial* approximation to the sign function and ensuring its *stability* are crucial.

Using matrix Chebyshev polynomials to approximate matrix functions is not new. Perhaps the most celebrated example is to approximate $\mathbf{S}^{-1}$ using polynomials on $\mathbf{S}$, used in the analysis of conjugate gradient [22]. Independent from this paper,[5] Han *et al.* [15] used Chebyshev polynomials to approximate the trace of the matrix sign function, i.e., $\mathbf{Tr}(\mathrm{sgn}(\mathbf{S}))$, which is similar but a different problem.[6] Also, they did not study the case when the matrix-vector multiplication oracle is only approximate (like we do in this paper), or the case when $\mathbf{S}$ has eigenvalues in the range $[-\gamma, \gamma]$.

[4]We anyways have included Krylov method in our empirical evaluation section and shall discuss its performance there, see for instance Remark 8.1.

[5]Their paper appeared online two months before us, and we became aware of their work in March 2017.

[6]In particular, their degree of the Chebyshev polynomial is $O\big(\gamma^{-1}(\log^2(1/\gamma) + \log(1/\gamma)\log(1/\varepsilon))\big)$ in the language of this paper; in contrast, we have degree $O\big(\gamma^{-1}\log(1/\gamma\varepsilon)\big)$.

**Roadmap.**

- In Section 2, we provide notions for this paper and basics for Chebyshev polynomials
- In Section 3, we put forward our formal definitions for approximate PCP and PCR, and show a reduction from approximate PCR to approximate PCP.
- In Section 4, we prove a general lemma regarding Chebyshev approximations outside $[-1,1]$.
- In Section 5, we design our polynomial approximation to $\mathrm{sgn}(x)$.
- In Section 6, we show how to stably compute any weighted sum of Chebyshev polynomials.
- In Section 7, we provide pseudocode and prove our main theorems regarding PCP and PCR.
- In Section 8, we provide empirical evaluations of our theory.

# 2 Preliminaries

We denote by $\mathbb{1}[e] \in \{0,1\}$ the indicator function for event $e$, by $\|v\|$ or $\|v\|_2$ the Euclidean norm of a vector $v$, by $\mathbf{M}^\dagger$ the Moore-Penrose pseudo-inverse of a symmetric matrix $\mathbf{M}$, and by $\|\mathbf{M}\|_2$ its spectral norm. We sometimes use $\vec{v}$ to emphasize that $v$ is a vector.

Given a symmetric $d \times d$ matrix $\mathbf{M}$ and any $f \colon \mathbb{R} \to \mathbb{R}$, $f(\mathbf{M})$ is the matrix function applied to $\mathbf{M}$, which is equal to $\mathbf{U}\mathsf{diag}\{f(D_1), \dots, f(D_d)\}\mathbf{U}^\top$ if $\mathbf{M} = \mathbf{U}\mathsf{diag}\{D_1, \dots, D_d\}\mathbf{U}^\top$ is its eigendecomposition.

Throughout the paper, matrix $\mathbf{A}$ is of dimension $d' \times d$. We denote by $\sigma_{\max}(\mathbf{A})$ the largest singular value of $\mathbf{A}$. Following the tradition of [13] and keeping the notations light, we assume without loss of generality that $\sigma_{\max}(\mathbf{A}) \leq 1$. We are interested in PCP and PCR problems with an eigenvalue threshold $\lambda \in (0,1)$.

Throughout the paper, we denote by $\lambda_1 \geq \cdots \geq \lambda_d \geq 0$ the eigenvalues of $\mathbf{A}^\top\mathbf{A}$, and by $\nu_1, \dots, \nu_d \in \mathbb{R}^d$ the eigenvectors of $\mathbf{A}^\top\mathbf{A}$ corresponding to $\lambda_1, \dots, \lambda_d$. We denote by $\mathbf{P}_\lambda$ the projection matrix $\mathbf{P}_\lambda \stackrel{\text{def}}{=} (\nu_1, \dots, \nu_j)(\nu_1, \dots, \nu_j)^\top$ where $j$ is the largest index satisfying $\lambda_j \geq \lambda$. In other words, $\mathbf{P}_\lambda$ is a projection matrix to the eigenvectors of $\mathbf{A}^\top\mathbf{A}$ with eigenvalues $\geq \lambda$.

**Definition 2.1.** *The principal component projection (PCP) of $\chi \in \mathbb{R}^d$ at threshold $\lambda$ is $\xi^* = \mathbf{P}_\lambda \chi$.*

**Definition 2.2.** *The principal component regression (PCR) of regressand $b \in \mathbb{R}^{d'}$ at threshold $\lambda$ is*

$$x^* = \arg\min_{y \in \mathbb{R}^d} \|\mathbf{A}\mathbf{P}_\lambda y - b\|_2 \quad \textit{or equivalently} \quad x^* = (\mathbf{A}^\top\mathbf{A})^\dagger \mathbf{P}_\lambda (\mathbf{A}^\top b) \enspace .$$

## 2.1 Ridge Regression

**Definition 2.3.** *A black-box algorithm* `ApxRidge`$(\mathcal{A}, \lambda, u)$ *is an $\varepsilon$-approximate ridge regression solver, if for every $u \in \mathbb{R}^d$, it satisfies* $\|\mathtt{ApxRidge}(\mathcal{A}, \lambda, u) - (\mathbf{A}^\top\mathbf{A} + \lambda\mathbf{I})^{-1}u\| \leq \varepsilon\|u\|$.[7]

Ridge regression is equivalent to solving well-conditioned linear systems, or minimizing strongly convex and smooth objectives $f(y) \stackrel{\text{def}}{=} \frac{1}{2} y^\top(\mathbf{A}^\top\mathbf{A} + \lambda\mathbf{I})y - u^\top y$.

*Remark* 2.4. There is huge literature on efficient algorithms solving ridge regression. Most notably,

(1) Conjugate gradient [22] or accelerated gradient descent [20] gives fastest full-gradient methods;

(2) SVRG [17] and its acceleration Katyusha [1] give the fastest stochastic-gradient method; and

(3) NUACDM [5] gives the fastest coordinate-descent method.

[7] In fact, throughout the paper, we only need `ApxRidge` to satisfy this property with high probability for each $u$.

The running time of (1) is $O(\mathsf{nnz}(\mathbf{A})\lambda^{-1/2}\log(1/\varepsilon))$ where $\mathsf{nnz}(\mathbf{A})$ is time to multiply $\mathbf{A}$ to any vector. The running times of (2) and (3) depend on structural properties of $\mathbf{A}$ and are always faster than (1).

Because the best complexity of ridge regression depends on the structural properties of $\mathbf{A}$, following Frostig *et al.*, we only compute our running time in terms of the "number of black-box calls" to a ridge regression solver.

## 2.2 Chebyshev Polynomials

**Definition 2.5.** *Chebyshev polynomials of 1st and 2nd kind are $\{\mathcal{T}_n(x)\}_{n\geq 0}$ and $\{\mathcal{U}_n(x)\}_{n\geq 0}$ where*

$$\mathcal{T}_0(x) \stackrel{\text{def}}{=} 1, \qquad \mathcal{T}_1(x) \stackrel{\text{def}}{=} x, \qquad \mathcal{T}_{n+1}(x) \stackrel{\text{def}}{=} 2x\cdot\mathcal{T}_n(x) - \mathcal{T}_{n-1}(x)$$
$$\mathcal{U}_0(x) \stackrel{\text{def}}{=} 1, \qquad \mathcal{U}_1(x) \stackrel{\text{def}}{=} 2x, \qquad \mathcal{U}_{n+1}(x) \stackrel{\text{def}}{=} 2x\cdot\mathcal{U}_n(x) - \mathcal{U}_{n-1}(x)$$

**Fact 2.6** ([23])**.** *It satisfies $\frac{d}{dx}\mathcal{T}_n(x) = n\mathcal{U}_{n-1}(x)$ for $n\geq 1$ and*

$$\forall n\geq 0: \quad \mathcal{T}_n(x) = \begin{cases} \cos(n\arccos(x)), & \text{if } |x|\leq 1; \\ \cosh(n\,\mathrm{arccosh}(x)), & \text{if } x\geq 1; \\ (-1)^n\cosh(n\,\mathrm{arccosh}(-x)), & \text{if } x\leq -1. \end{cases}$$

*In particular, when $x\geq 1$, $\mathcal{T}_n(x) = \frac{1}{2}\big[\big(x-\sqrt{x^2-1}\big)^n + \big(x+\sqrt{x^2-1}\big)^n\big]$ and $\mathcal{U}_n(x) = \frac{1}{2\sqrt{x^2-1}}\big[\big(x+\sqrt{x^2-1}\big)^{n+1} - \big(x-\sqrt{x^2-1}\big)^{n+1}\big]$.*

$$\mathcal{T}_n(x) = \frac{1}{2}\big[\big(x-\sqrt{x^2-1}\big)^n + \big(x+\sqrt{x^2-1}\big)^n\big]$$
$$\mathcal{U}_n(x) = \frac{1}{2\sqrt{x^2-1}}\big[\big(x+\sqrt{x^2-1}\big)^{n+1} - \big(x-\sqrt{x^2-1}\big)^{n+1}\big]$$

**Definition 2.7.** *For function $f(x)$ whose domain contains $[-1,1]$, its degree-$n$ Chebyshev truncated series and degree-$n$ Chebyshev interpolation are respectively*

$$p_n(x) \stackrel{\text{def}}{=} \sum_{k=0}^{n} a_k\mathcal{T}_k(x) \quad \text{and} \quad q_n(x) \stackrel{\text{def}}{=} \sum_{k=0}^{n} c_k\mathcal{T}_k(x) \ ,$$

$$\text{where } a_k \stackrel{\text{def}}{=} \frac{2-\mathbb{1}[k=0]}{\pi}\int_{-1}^{1}\frac{f(x)\mathcal{T}_k(x)}{\sqrt{1-x^2}}dx \quad \text{and} \quad c_k \stackrel{\text{def}}{=} \frac{2-\mathbb{1}[k=0]}{n+1}\sum_{j=0}^{n} f\big(x_j\big)\mathcal{T}_k\big(x_j\big) \ .$$

*Above, $x_j \stackrel{\text{def}}{=} \cos\big(\frac{(j+0.5)\pi}{n+1}\big) \in [-1,1]$ is the $j$-th Chebyshev point of order $n$.*

The following lemma is known as the aliasing formula for Chebyshev coefficients:

**Lemma 2.8** (cf. Theorem 4.2 of [23])**.** *Let $f$ be Lipschitz continuous on $[-1,1]$ and $\{a_k\},\{c_k\}$ be defined in Def. 2.7, then*

$$c_0 = a_0 + a_{2n} + a_{4n} + \dots \quad , \quad c_n = a_n + a_{3n} + a_{5n} + \dots \quad , \text{ and}$$

$$\forall k\in\{1,2,\dots,n-1\}: \quad c_k = a_k + (a_{k+2n} + a_{k+4n} + \dots) + (a_{-k+2n} + a_{-k+4n} + \dots)$$

**Definition 2.9.** *For every $\rho > 0$, let $\mathcal{E}_\rho$ be the ellipse $\mathcal{E}$ of foci $\pm 1$ with major radius $1+\rho$. (This is also known as Bernstein ellipse with parameter $1+\rho+\sqrt{2\rho+\rho^2}$.)*

The following lemma is the main theory regarding Chebyshev approximation:

**Lemma 2.10** (cf. Theorem 8.1 and 8.2 of [23]). *Suppose $f(z)$ is analytic on $\mathcal{E}_\rho$ and $|f(z)| \le M$ on $\mathcal{E}_\rho$. Let $p_n(x)$ and $q_n(x)$ be the degree-$n$ Chebyshev truncated series and Chebyshev interpolation of $f(x)$ on $[-1,1]$. Then,*

- $\max_{x\in[-1,1]} |f(x) - p_n(x)| \le \frac{2M}{\rho+\sqrt{2\rho+\rho^2}}\big(1+\rho+\sqrt{2\rho+\rho^2}\big)^{-n}$;
- $\max_{x\in[-1,1]} |f(x) - q_n(x)| \le \frac{4M}{\rho+\sqrt{2\rho+\rho^2}}\big(1+\rho+\sqrt{2\rho+\rho^2}\big)^{-n}$.
- $|a_0| \le M$ *and* $|a_k| \le 2M\big(1+\rho+\sqrt{2\rho+\rho^2}\big)^{-k}$ *for* $k \ge 1$.

# 3 Approximate PCP and PCR

We formalize our notions of approximation for PCP and PCR, and provide a reduction from PCR to PCP.

## 3.1 Our Notions of Approximation

Recall that Frostig *et al.* [13] work only with matrices $\mathbf{A}$ that satisfy the eigengap assumption, that is, $\mathbf{A}$ has no singular value in the range $[\sqrt{\lambda(1-\gamma)}, \sqrt{\lambda(1+\gamma)}]$. Their approximation guarantees are very straightforward:

- an output $\xi$ is $\varepsilon$-approximate for PCP on vector $\chi$ if $\|\xi - \xi^*\| \le \varepsilon\|\chi\|$;
- an output $x$ is $\varepsilon$-approximate for PCR with regressand $b$ if $\|x - x^*\| \le \varepsilon\|b\|$.

Unfortunately, these notions are too strong and impossible to satisfy for matrices that do not have a large eigengap around the projection threshold $\lambda$.

In this paper we propose the following more general (but yet very meaningful) approximation notions.

**Definition 3.1.** *An algorithm $\mathcal{B}(\chi)$ is $(\gamma,\varepsilon)$-approximate PCP for threshold $\lambda$, if for every $\chi \in \mathbb{R}^d$*

1. $\big\|\mathbf{P}_{(1+\gamma)\lambda}\big(\mathcal{B}(\chi) - \chi\big)\big\| \le \varepsilon\|\chi\|$.
2. $\big\|(\mathbf{I} - \mathbf{P}_{(1-\gamma)\lambda})\mathcal{B}(\chi)\big\| \le \varepsilon\|\chi\|$.
3. $\forall i$ *such that* $\lambda_i \in \big[(1-\gamma)\lambda, (1+\gamma)\lambda\big]$, *it satisfies* $|\langle \nu_i, \mathcal{B}(\chi) - \chi\rangle| \le |\langle \nu_i, \chi\rangle| + \varepsilon\|\chi\|$.

Intuitively, the first property above states that, if projected to the eigenspace with eigenvalues above $(1+\gamma)\lambda$, then $\mathcal{B}(\chi)$ and $\chi$ are almost identical; the second property states that, if projected to the eigenspace with eigenvalues below $(1-\gamma)\lambda$, then $\mathcal{B}(\chi)$ is almost zero; and the third property states that, for each eigenvector $\nu_i$ with eigenvalue in the range $[(1-\gamma)\lambda, (1+\gamma)\lambda]$, the projection $\langle \nu_i, \mathcal{B}(\chi)\rangle$ must be between 0 and $\langle \nu_i, \chi\rangle$ (but up to an error $\varepsilon\|\chi\|$).

Naturally, $\mathbf{P}_\lambda(\chi)$ itself is a $(0,0)$-approximate PCP.

We propose the following notion for approximate PCR:

**Definition 3.2.** *An algorithm $\mathcal{C}(b)$ is $(\gamma,\varepsilon)$-approximate PCR for threshold $\lambda$, if for every $b \in \mathbb{R}^{d'}$*

1. $\big\|(\mathbf{I} - \mathbf{P}_{(1-\gamma)\lambda})\mathcal{C}(b)\big\| \le \varepsilon\|b\|$.
2. $\|\mathbf{A}\mathcal{C}(b) - b\| \le \|\mathbf{A}x^* - b\| + \varepsilon\|b\|$.

*where $x^* = (\mathbf{A}^\top\mathbf{A})^\dagger \mathbf{P}_{(1+\gamma)\lambda}\mathbf{A}^\top b$ is the exact PCR solution for threshold $(1+\gamma)\lambda$.*

The first notion states that the output $x = \mathcal{C}(b)$ has nearly no correlation with eigenvectors below threshold $(1-\gamma)\lambda$; and the second states that the regression error should be nearly optimal with respect to the exact PCR solution but at a different threshold $(1+\gamma)\lambda$.

**Relationship to Frostig *et al.*** Under eigengap assumption, our notions are equivalent to Frostig *et al.*:

**Fact 3.3.** *If* $\mathbf{A}$ *has no singular value in* $[\sqrt{\lambda(1-\gamma)}, \sqrt{\lambda(1+\gamma)}]$, *then*

- *Def. 3.1 is equivalent to* $\|\mathcal{B}(\chi) - \mathbf{P}_\lambda(\chi)\| \le O(\varepsilon)\|\chi\|$.
- *Def. 3.2 implies* $\|\mathcal{C}(\chi) - x^*\| \le O(\varepsilon/\lambda)\|b\|$ *and* $\|\mathcal{C}(\chi) - x^*\| \le O(\varepsilon)\|b\|$ *implies Def. 3.2.*

*Above,* $x^* = (\mathbf{A}^\top\mathbf{A})^\dagger \mathbf{P}_\lambda \mathbf{A}^\top b$ *is the exact PCR solution.*

## 3.2 Reductions from PCR to PCP

If the PCP solution $\xi = \mathbf{P}_\lambda(\mathbf{A}^\top b)$ is computed exactly, then by definition one can compute $(\mathbf{A}^\top\mathbf{A})^\dagger\xi$ which gives a solution to PCR by solving a linear system. However, as pointed by Frostig *et al.* [13], this computation is problematic if $\xi$ is only approximate. The following approach has been proposed to improve its accuracy by Frostig *et al.*

- "compute $p((\mathbf{A}^\top\mathbf{A} + \lambda\mathbf{I})^{-1})\xi$ where $p(x)$ is a polynomial that approximates function $\frac{x}{1-\lambda x}$."

This is a good approximation to $(\mathbf{A}^\top\mathbf{A})^\dagger\xi$ because the composition of functions $\frac{x}{1-\lambda x}$ and $\frac{1}{1+\lambda x}$ is exactly $x^{-1}$. Frostig *et al.* picked $p(x) = p_m(x) = \sum_{t=1}^m \lambda^{t-1}x^t$ which is a truncated Taylor series, and used the following procedure to compute $s_m \approx p_m((\mathbf{A}^\top\mathbf{A} + \lambda\mathbf{I})^{-1})\xi$:

$$s_0 = \mathcal{B}(\mathbf{A}^\top b), \quad s_1 = \texttt{ApxRidge}(\mathbf{A}, \lambda, s_0), \quad \forall k \ge 1\colon s_{k+1} = s_1 + \lambda \cdot \texttt{ApxRidge}(\mathbf{A}, \lambda, s_k) \ . \tag{3.1}$$

Above, $\mathcal{B}$ is an approximate PCP solver and `ApxRidge` is an approximate ridge regression solver.

Under the *eigengap assumption*, Frostig *et al.* [13] showed that

**Lemma 3.4** (PCR-to-PCP)**.** *For fixed* $\lambda, \gamma, \varepsilon \in (0,1)$, *let* $\mathbf{A}$ *be a matrix whose singular values lie in* $\big[0, \sqrt{(1-\gamma)\lambda}\big] \cup \big[\sqrt{(1-\gamma)\lambda}, 1\big]$. *Let* `ApxRidge` *be any* $O(\frac{\varepsilon}{m^2})$*-approximate ridge regression solver, and let* $\mathcal{B}$ *be any* $(\gamma, O(\frac{\varepsilon\lambda}{m^2}))$*-approximate PCP solver*[8]. *Then, procedure (3.1) satisfies*

$$\|s_m - (\mathbf{A}^\top\mathbf{A})^\dagger \mathbf{P}_\lambda \mathbf{A}^\top b\| \le \varepsilon\|b\| \quad \textit{if} \quad m = \Theta(\log(1/\varepsilon\gamma)) \ .$$

Unfortunately, the above lemma *does not hold* without eigengap assumption. In this paper, we fix this issue by proving the following analogous lemma:

**Lemma 3.5** (gap free PCR-to-PCP)**.** *For fixed* $\lambda, \varepsilon \in (0,1)$ *and* $\gamma \in (0, 2/3]$, *let* $\mathbf{A}$ *be a matrix whose singular values are no more than* 1. *Let* `ApxRidge` *be any* $O(\frac{\varepsilon}{m^2})$*-approximate ridge regression solver, and* $\mathcal{B}$ *be any* $(\gamma, O(\frac{\varepsilon\lambda}{m^2}))$*-approximate PCP solver. Then, procedure (3.1) satisfies,*

$$\left\{ \begin{array}{c} \|(\mathbf{I} - \mathbf{P}_{(1-\gamma)\lambda})s_m\| \le \varepsilon\|b\| \quad , \textit{ and} \\ \|\mathbf{A}s_m - b\| \le \|\mathbf{A}(\mathbf{A}^\top\mathbf{A})^\dagger \mathbf{P}_{(1+\gamma)\lambda}\mathbf{A}^\top b - b\| + \varepsilon\|b\| \end{array} \right\} \quad \textit{if} \quad m = \Theta(\log(1/\varepsilon\gamma))$$

Note that the conclusion of this lemma exactly corresponds to the two properties in our Def. 3.2. The proof of Lemma 3.5 is not hard, but requires a very careful case analysis by decomposing vectors $b$ and each $s_k$ into three components, each corresponding to eigenvalues of $\mathbf{A}^\top\mathbf{A}$ in the range $[0, (1-\gamma)\lambda]$, $[(1-\gamma)\lambda, (1+\gamma)\lambda]$ and $[(1+\gamma)\lambda, 1]$. We defer the details to Appendix A.

[8] Recall from Fact 3.3 that this requirement is equivalent to saying that $\|\mathcal{B}(\chi) - \mathbf{P}_\lambda\chi\| \le O(\frac{\varepsilon\sqrt{\lambda}}{m^2})\|\chi\|$.

# 4 Property of Chebyshev Approximation Outside $[-1,1]$

Classical Chebyshev approximation theory (such as Lemma 2.10) only talks about the behaviors of $p_n(x)$ or $g_n(x)$ on interval $[-1,1]$. However, for the purpose of this paper, we must also bound its value for $x>1$. We prove the following general lemma in Appendix B, and believe it could be of independent interest: (we denote by $f^{(k)}(x)$ the $k$-th derivative of $f$ at $x$)

**Lemma 4.1.** *Suppose $f(z)$ is analytic on $\mathcal{E}_\rho$ and for every $k\geq 0$, $f^{(k)}(0)\geq 0$. Then, for every $n\in\mathbb{N}$, letting $p_n(x)$ and $q_n(x)$ be be the degree-$n$ Chebyshev truncated series and Chebyshev interpolation of $f(x)$, we have*

$$\forall y\in[0,\rho]:\quad 0\leq p_n(1+y), q_n(1+y)\leq f(1+y)\ .$$

# 5 Our Polynomial Approximation of $\operatorname{sgn}(x)$

For fixed $\kappa\in(0,1]$, we consider the degree-$n$ Chebyshev interpolation $q_n(x)=\sum_{k=0}^n c_k\mathcal{T}_k(x)$ of the function $f(x)=\left(\frac{1+\kappa-x}{2}\right)^{-1/2}$ on $[-1,1]$. Def. 2.7 tells us that

$$c_k \stackrel{\text{def}}{=} \frac{2-\mathbb{1}[k=0]}{n+1}\sum_{j=0}^{n}\left(\sqrt{2}\cos\left(\frac{k(j+0.5)\pi}{n+1}\right)\right)\left(1+\kappa-\cos\left(\frac{(j+0.5)\pi}{n+1}\right)\right)^{-1/2} .$$

Our final polynomial to approximate $\operatorname{sgn}(x)$ is therefore

$$g_n(x)=x\cdot q_n(1+\kappa-2x^2)\quad\text{and}\quad\deg(g_n(x))=2n+1\ .$$

We prove the following theorem in this section:

**Theorem 5.1.** *For every $\alpha\in(0,1],\varepsilon\in(0,1/2)$, choosing $\kappa=2\alpha^2$, our function $g_n(x)\stackrel{\text{def}}{=} x\cdot q_n(1+\kappa-2x^2)$ satisfies that as long as $n\geq\frac{1}{\sqrt{2}\alpha}\log\frac{3}{\varepsilon\alpha^2}$, then (see also Figure 1)*

- *$|g_n(x)-\operatorname{sgn}(x)|\leq\varepsilon$ for every $x\in[-1,\alpha]\cup[\alpha,1]$.*
- *$g_n(x)\in[0,1]$ for every $x\in[0,\alpha]$ and $g_n(x)\in[-1,0]$ for every $x\in[-\alpha,0]$.*

Note that our degree $n=O\big(\alpha^{-1}\log(1/\alpha\varepsilon)\big)$ is near-optimal, because the minimum degree for a polynomial to satisfy even only the first item is $\Theta\big(\alpha^{-1}\log(1/\varepsilon)\big)$ [9, 10]. However, the results of [9, 10] are not constructive, and thus may not lead to stable matrix polynomials.

We prove Theorem 5.1 by first establishing two simple lemmas. The following lemma is a consequence of Lemma 2.10:

**Lemma 5.2.** *For every $\varepsilon\in(0,1/2)$ and $\kappa\in(0,1]$, if $n\geq\frac{1}{\sqrt{\kappa}}\left(\log\frac{1}{\kappa}+\log\frac{4}{\varepsilon}\right)$ then*

$$\forall x\in[-1,1],\quad |f(x)-q_n(x)|\leq\varepsilon\ .$$

*Proof of Lemma 5.2.* Denoting by $f(z)=\left(\frac{1+\kappa-z}{2}\right)^{-0.5}$, we know that $f(z)$ is analytic on ellipse $\mathcal{E}_\rho$ with $\rho=\kappa/2$, and it satisfies $|f(z)|\leq\sqrt{2/\kappa}$ in $\mathcal{E}_\rho$. Applying Lemma 2.10, we know that when $n\geq\frac{1}{\sqrt{\kappa}}\left(\log\frac{1}{\kappa}+\log\frac{4}{\varepsilon}\right)$ it satisfies $|f(x)-q_n(x)|\leq\varepsilon$. □

The next lemma an immediate consequence of our Lemma 4.1 with $f(z)=\left(\frac{1+\kappa-z}{2}\right)^{-0.5}$:

**Lemma 5.3.** *For every $\varepsilon\in(0,1/2),\kappa\in(0,1],n\in\mathbb{N}$, and $x\in[0,\kappa]$, we have*

$$0\leq q_n(1+x)\leq\left(\frac{\kappa-x}{2}\right)^{-1/2}\ .$$

*Proof of Theorem 5.1.* We are now ready to prove Theorem 5.1.

- When $x \in [-1, \alpha] \cup [\alpha, 1]$, it satisfies $1 + \kappa - 2x^2 \in [-1, 1]$. Therefore, applying Lemma 5.2 we have whenever $n \geq \frac{1}{\sqrt{\kappa}} \log \frac{6}{\varepsilon\kappa} = \frac{1}{\sqrt{2}\alpha} \log \frac{3}{\varepsilon\alpha^2}$ it satisfies $|f(1+\kappa-2x^2) - q_n(1+\kappa-2x^2)|_\infty \leq \varepsilon$. This further implies

  $$|g_n(x) - \mathrm{sgn}(x)| = |xq_n(1+\kappa-2x^2) - xf(1+\kappa-2x^2)| \leq |x||f(1+\kappa-2x^2) - q_n(1+\kappa-2x^2)| \leq \varepsilon \ .$$

- When $|x| \leq \alpha$, it satisfies $1 + \kappa - 2x^2 \in [1, 1+\kappa]$. Applying Lemma 5.3 we have

  $$\forall x \in [0, \alpha]: \quad 0 \leq g_n(x) = x \cdot q_n(1+\kappa-2x^2) \leq x \cdot (x^2)^{-1/2} = 1$$

  and similarly for $x \in [-\alpha, 0]$ it satisfies $0 \geq g_n(x) \geq -1$. □

**A Bound on Chebyshev Coefficients.** We also give an upper bound to the coefficients of polynomial $q_n(x)$. Its proof can be found in Appendix C, and this upper bound shall be used in our final stability analysis.

**Lemma 5.4** (coefficients of $q_n$)**.** *Let $q_n(x) = \sum_{k=0}^{n} c_k \mathcal{T}_k(x)$ be the degree-n Chebyshev interpolation of $f(x) = \left(\frac{1+\kappa-x}{2}\right)^{-1/2}$ on $[-1, 1]$. Then,*

$$\forall i \in \{0, 1, \ldots, n\}: \quad |c_i| \leq \frac{e\sqrt{32(i+1)}}{\kappa} \left(1 + \kappa + \sqrt{2\kappa + \kappa^2}\right)^{-i}$$

# 6 Stable Computation of Matrix Chebyshev Polynomials

In this section we show that any polynomial that is a weighted summation of Chebyshev polynomials with bounded coefficients, can be stably computed when applied to matrices with approximate computations. We achieve so by first generalizing Clenshaw's backward method to matrix case in Section 6.1 in order to compute a matrix variant of Chebyshev sum, and then analyze its stability in Section 6.2 with the help from Elloit's forward-backward transformation [8].

*Remark* 6.1. We wish to point out that although Chebyshev polynomials are known to be stable under error when computed on *scalars* [14], it is not immediately clear why it holds also for matrices. Recall that Chebyshev polynomials satisfy $\mathcal{T}_{n+1}(x) = 2x\mathcal{T}_n(x) - \mathcal{T}_{n-1}(x)$. In the matrix case, we have $\mathcal{T}_{n+1}(\mathbf{M})\chi = 2\mathbf{M}\mathcal{T}_n(\mathbf{M})\chi - \mathcal{T}_{n-1}(\mathbf{M})\chi$ where $\chi \in \mathbb{R}^d$ is a vector. If we analyzed this formula coordinate by coordinate, error could blow up by a factor $d$ per iteration.

In addition, we need to ensure that the stability theorem holds for matrices $\mathbf{M}$ with eigenvalues that can exceed 1. This is not standard because Chebyshev polynomials are typically analyzed only on domain $[-1, 1]$.

## 6.1 Clenshaw's Method in Matrix Form

In the scalar case, Clenshaw's method (sometimes referred to as backward recurrence) is one of the most widely used implementations for Chebyshev polynomials. We now generalize it to matrices.

Consider any computation of the form

$$\vec{s}_N \stackrel{\text{def}}{=} \sum_{k=0}^{N} \mathcal{T}_k(\mathbf{M})\vec{c}_k \in \mathbb{R}^d \quad \text{where } \mathbf{M} \in \mathbb{R}^{d\times d} \text{ is symmetric and each } \vec{c}_k \text{ is in } \mathbb{R}^d \ . \tag{6.1}$$

(Note that for PCP and PCR purposes, we it suffices to consider $\vec{c}_k = c'_k\chi$ where $c'_k \in \mathbb{R}$ is a scalar and $\chi \in \mathbb{R}^d$ is a fixed vector for all $k$. However, we need to work on this more general form for our stability analysis.)

Vector $s_N$ can be computed using the following procedure:

**Lemma 6.2** (backward recurrence)**.** $\vec{s}_N = \vec{b}_0 - \mathbf{M}\vec{b}_1$ *where*

$$\vec{b}_{N+1} \stackrel{\text{def}}{=} \vec{0}, \quad \vec{b}_N \stackrel{\text{def}}{=} \vec{c}_N, \quad \text{and} \quad \forall r \in \{N-1, \dots, 0\} \colon \vec{b}_r \stackrel{\text{def}}{=} 2\mathbf{M}\vec{b}_{r+1} - \vec{b}_{r+2} + \vec{c}_r \in \mathbb{R}^d \ .$$

### 6.2 Inexact Clenshaw's Method in Matrix Form

We show that, if implemented using the backward recurrence formula, the Chebyshev sum of (6.1) can be stably computed. We define the following model to capture the error with respect to matrix-vector multiplications.

**Definition 6.3** (inexact backward recurrence)**.** *Let $\mathcal{M}$ be an approximate algorithm that satisfies $\|\mathcal{M}(u) - \mathbf{M}u\|_2 \le \varepsilon \|u\|_2$ for every $u \in \mathbb{R}^d$. Then, define inexact backward recurrence to be*

$$\widehat{b}_{N+1} \stackrel{\text{def}}{=} 0, \quad \widehat{b}_N \stackrel{\text{def}}{=} \vec{c}_N, \quad \text{and} \quad \forall r \in \{N-1, \dots, 0\} \colon \widehat{b}_r \stackrel{\text{def}}{=} 2\mathcal{M}\big(\widehat{b}_{r+1}\big) - \widehat{b}_{r+2} + \vec{c}_r \in \mathbb{R}^d \ ,$$

*and define the output as $\widehat{s}_N \stackrel{\text{def}}{=} \widehat{b}_0 - \mathcal{M}(\widehat{b}_1)$.*

The following theorem gives an error analysis to our inexact backward recurrence. We prove it in Appendix D.1, and the main idea of our proof is to convert each error vector of a recursion of the backward procedure into an error vector corresponding to some original $\vec{c}_k$.

**Theorem 6.4** (stable Chebyshev sum)**.** *For every $N \in \mathbb{N}^*$, suppose the eigenvalues of $\mathbf{M}$ are in $[a, b]$ and suppose there are parameters $C_U \ge 1, C_T \ge 1, \rho \ge 1, C_c \ge 0$ satisfying*

$$\forall k \in \{0, 1, \dots, N\} \colon \Big\{ \rho^k \|\vec{c}_k\| \le C_c \quad \bigwedge \quad \forall x \in [a, b] \colon \quad |\mathcal{T}_k(x)| \le C_T \rho^k \text{ and } |\mathcal{U}_k(x)| \le C_U \rho^k \Big\} \ .$$

*Then, if the inexact backward recurrence in Def. 6.3 is applied with $\varepsilon \le \frac{1}{4NC_U}$, we have*

$$\|\widehat{s}_N - \vec{s}_N\| \le \varepsilon \cdot 2(1 + 2NC_T)NC_U C_c \ .$$

## 7 Algorithms and Main Theorems for PCP and PCR

We are now ready to state our main theorems for PCP and PCR. We first note a simple fact:

**Fact 7.1.** $(\mathbf{P}_\lambda)\chi = \frac{\mathbf{I} + \mathrm{sgn}(\mathbf{S})}{2}$ *where* $\mathbf{S} \stackrel{\text{def}}{=} 2(\mathbf{A}^\top\mathbf{A} + \lambda\mathbf{I})^{-1}\mathbf{A}^\top\mathbf{A} - \mathbf{I} = (\mathbf{A}^\top\mathbf{A} + \lambda\mathbf{I})^{-1}(\mathbf{A}^\top\mathbf{A} - \lambda\mathbf{I})$.

In other words, for every vector $\chi \in \mathbb{R}^d$, the exact PCP solution $\mathbf{P}_\lambda(\chi)$ is the same as computing $(\mathbf{P}_\lambda)\chi = \frac{\mathbf{I} + \mathrm{sgn}(\mathbf{S})}{2}\chi$. Thus, we can use our polynomial $g_n(x)$ introduced in Section 5 and compute $g_n(\mathbf{S})\chi \approx \mathrm{sgn}(\mathbf{S})\chi$. Finally, in order to compute $g_n(\mathbf{S})$, we need to multiply $\mathbf{S}$ to $\deg(g_n)$ vectors; whenever we do so, we call perform ridge regression once.

### 7.1 Our Pseudo Codes

First of all, we can approximately compute $\mathbf{S}\chi$ for an arbitrary $\chi \in \mathbb{R}^d$. This simply uses one oracle call to ridge regression, see Algorithm 1.

Next, since we are interested in $(\gamma, \varepsilon)$-approximate PCP, we want $g_n(x)$ to be close to $\mathrm{sgn}(x)$ on all eigenvalues of $\mathbf{A}^\top\mathbf{A}$ that are outside $[(1-\gamma)\lambda, (1+\gamma)\lambda]$, or equivalently all eigenvalues of $\mathbf{S}$ outside the range

$$\Big[ -\frac{(1+\gamma) - 1}{1 + (1+\gamma)}, \frac{1 - (1-\gamma)}{1 + (1-\gamma)} \Big] \ .$$

**Algorithm 1** $\texttt{MultS}(\mathbf{A}, \lambda, \chi)$

**Input:** $\mathbf{A} \in \mathbb{R}^{d' \times d}$; $\lambda > 0$; $\chi \in \mathbb{R}^d$.
**Output:** a vector that approximately equals $\mathbf{S}\chi = (\mathbf{A}^\top \mathbf{A} + \lambda \mathbf{I})^{-1}(\mathbf{A}^\top \mathbf{A} - \lambda \mathbf{I})\chi$

**return** $\texttt{ApxRidge}(\mathbf{A}, \lambda, \mathbf{A}^\top \mathbf{A}\chi - \lambda\chi)$.

Since this new interval contains $[-\alpha, \alpha]$ for $\alpha \stackrel{\text{def}}{=} \gamma/(2+\gamma) = \gamma/2 - O(\gamma^2)$, we can apply Theorem 5.1, which gives us a polynomial $g_n(x) = x \cdot q_n(1 + \kappa - 2x^2)$ where $\kappa = 2\alpha^2 = 2(\gamma/(2+\gamma))^2$. We use (inexact) backward recurrence —see Lemma 6.2— to compute the Chebyshev interpolation polynomial $u \leftarrow q_n\big((1+\kappa)\mathbf{I} - 2\mathbf{S}^2\big)\chi$. Our final output for approximate PCP is simply $\frac{\mathbf{S}u+\chi}{2}$ because $\mathbf{P}_\lambda \approx \frac{\mathbf{S}g_n((1+\kappa)-2\mathbf{S}^2)+\mathbf{I}}{2}$. We summarize this algorithm as $\texttt{QuickPCP}(\mathbf{A}, \chi, \lambda, \gamma, n)$ in Algorithm 2.

**Algorithm 2** $\texttt{QuickPCP}(\mathbf{A}, \chi, \lambda, \gamma, n)$

**Input:** $\mathbf{A} \in \mathbb{R}^{d' \times d}$ data matrix satisfying $\sigma_{\max}(\mathbf{A}) \le 1$; $\chi \in \mathbb{R}^d$, vector to project;
$\lambda > 0$, eigenvalue threshold for PCP; $\gamma \in (0, 2/3]$, PCP approximation ratio.
$n$, number of iterations ◇ *one can also ignore $\gamma$ and set $\gamma = 0$, see Remark 7.5*
**Output:** a vector $\xi \in \mathbb{R}^d$ satisfying $\xi \approx \mathbf{P}_\lambda(\chi)$.

$\gamma \leftarrow \max\{\gamma, \frac{\log(n)}{n}\}$ ◇ *if $\gamma$ to small, work in a $\gamma$-free regime, see Remark 7.5*
$\kappa \leftarrow 2\big(\gamma/(2+\gamma)\big)^2$ ◇ *recall $\kappa = 2\alpha^2 = 2(\gamma/(2+\gamma))^2$ in our analysis*
Define $c_k \stackrel{\text{def}}{=} \frac{2 - \mathbb{1}[k=0]}{n+1} \sum_{j=0}^{n} \Big(\sqrt{2}\cos\big(\frac{k(j+0.5)\pi}{n+1}\big)\Big)\Big(1 + \kappa - \cos\big(\frac{(j+0.5)\pi}{n+1}\big)\Big)^{-1/2}$ ◇ *coefficients for $q_n(x)$*
$b_{n+1} \leftarrow \vec{0}$, $b_n \leftarrow c_n \cdot \chi$
**for** $r \leftarrow n-1$ **to** $0$ **do**
  $w \leftarrow (1+\kappa)b_{r+1} - 2 \cdot \texttt{MultS}(\mathbf{A}, \lambda, \texttt{MultS}(\mathbf{A}, \lambda, b_{r+1}))$; ◇ $w \approx ((1+\kappa)\mathbf{I} - \mathbf{S}^2)b_{r+1}$
  $b_r \leftarrow 2w - b_{r+2} + c_r \cdot \chi$
**end for**
$u \leftarrow \texttt{MultS}(\mathbf{A}, \lambda, b_0 - w)$; ◇ $u \approx \mathbf{S}(g_n((1+\kappa)\mathbf{I} - \mathbf{S}^2))\chi \approx \mathrm{sgn}(\mathbf{S})\chi$
**return** $\frac{1}{2}u + \frac{1}{2}\chi$ ◇ *output* $\approx \frac{\mathrm{sgn}(\mathbf{S})+\mathbf{I}}{2}\chi$

Finally, we apply the PCR-to-PCP reduction (see Section 3) to derive a solution for PCR from an approximate solution for PCP. See $\texttt{QuickPCR}(\mathbf{A}, b, \lambda, \gamma, n, m)$ in Algorithm 3.

**Algorithm 3** $\texttt{QuickPCR}(\mathbf{A}, b, \lambda, \gamma, n, m)$

**Input:** $\mathbf{A}, \lambda, \gamma, n$ the same as $\texttt{QuickPCP}$; $b \in \mathbb{R}^{d'}$ is the regressand vector; $m$ is the number of iterations for PCR. ◇ *choosing $m = 10$ it sufficient for practical purposes*
**Output:** a vector $x \in \mathbb{R}^d$ that solves approximate PCR.

$v \leftarrow \texttt{QuickPCP}(\mathbf{A}, \mathbf{A}^\top b, \lambda, \gamma, n)$, $s \leftarrow v$, $s_1 \leftarrow \texttt{ApxRidge}(\mathbf{A}, \lambda, v)$;
**for** $r \leftarrow 1$ **to** $m$ **do**
  $s \leftarrow \lambda \cdot \texttt{ApxRidge}(\mathbf{A}, \lambda, s) + s_1$;
**return** s

**Fact 7.2.** $\texttt{QuickPCP}$ *calls ridge regression $2n+1$ times and* $\texttt{QuickPCR}$ *calls it $2n+m+2$ times.*

### 7.2 Our Main Theorems

We first state our main theorem under the eigengap assumption, in order to provide a direct comparison to that of Frostig *et al.* [13].

**Theorem 7.3** (eigengap assumption). *Given $\mathbf{A} \in \mathbb{R}^{d' \times d}$ and $\lambda, \gamma \in (0,1)$, assume that the singular values of $\mathbf{A}$ are in the range $[0, \sqrt{(1-\gamma)\lambda}] \cup [\sqrt{(1+\gamma)\lambda}, 1]$. Given $\chi \in \mathbb{R}^d$ and $b \in \mathbb{R}^{d'}$, denote by $\xi^* = \mathbf{P}_\lambda \chi$ and $x^* = (\mathbf{A}^\top \mathbf{A})^{-1} \mathbf{P}_\lambda \mathbf{A}^\top b$ the exact PCP and PCR solutions. If* `ApxRidge` *is an $\varepsilon'$-approximate ridge regression solver, then*

$$\text{the output} \quad \xi \leftarrow \texttt{QuickPCP}(\mathbf{A}, \chi, \lambda, \gamma, n) \quad \text{satisfies} \quad \|\xi^* - \xi\| \le \varepsilon \|\chi\|$$

*if $n = \Theta\big(\gamma^{-1} \log \frac{1}{\gamma\varepsilon}\big)$ and $\log(1/\varepsilon') = \Theta\big(\log \frac{1}{\gamma\varepsilon}\big)$;*

$$\text{the output} \quad x \leftarrow \texttt{QuickPCR}(\mathbf{A}, b, \lambda, \gamma, n, m) \quad \text{satisfies} \quad \|x - x^*\| \le \varepsilon \|b\|$$

*if $n = \Theta\big(\gamma^{-1} \log \frac{1}{\gamma\lambda\varepsilon}\big)$, $m = \Theta\big(\log \frac{1}{\gamma\varepsilon}\big)$ and $\log(1/\varepsilon') = \Theta\big(\log \frac{1}{\gamma\lambda\varepsilon}\big)$.*

In contrast, the number of ridge-regression oracle calls was $\Theta(\gamma^{-2} \log \frac{1}{\gamma\varepsilon})$ for PCP and $\Theta(\gamma^{-2} \log \frac{1}{\gamma\lambda\varepsilon})$ for PCR in [13]. We include the proof of Theorem 7.3 in Appendix E.1.

Next we state our stronger theorem without the eigengap assumption.

**Theorem 7.4** (gap-free). *Given $\mathbf{A} \in \mathbb{R}^{d' \times d}$, $\lambda \in (0,1)$, and $\gamma \in (0, 2/3]$, assume that $\|\mathbf{A}\|_2 \le 1$. Given $\chi \in \mathbb{R}^d$ and $b \in \mathbb{R}^{d'}$, and suppose* `ApxRidge` *is an $\varepsilon'$-approximate ridge regression solver, then*

- `QuickPCP` *outputs $\xi$ that is $(\gamma, \varepsilon)$-approximate PCP with $O\big(\gamma^{-1} \log \frac{1}{\gamma\varepsilon}\big)$ oracle calls to* `ApxRidge` *as long as $\log(1/\varepsilon') = \Theta\big(\log \frac{1}{\gamma\varepsilon}\big)$.*
- `QuickPCR` *outputs $x$ that is $(\gamma, \varepsilon)$-approximate PCR with $O\big(\gamma^{-1} \log \frac{1}{\gamma\lambda\varepsilon}\big)$ oracle calls to* `ApxRidge` *as long as $e\log(1/\varepsilon') = \Theta\big(\log \frac{1}{\gamma\lambda\varepsilon}\big)$.*

We make a final remark here regarding the practical usage of `QuickPCP` and `QuickPCR`.

*Remark* 7.5. Since our theory is for $(\gamma, \varepsilon)$-approximations that have two parameters, the user in principle has to feed in both $\gamma$ and $n$ (in addition to other default inputs such as $\mathbf{A}$, $b$ and $\lambda$). In practice, however, it is usually sufficient to obtain $(\varepsilon, \varepsilon)$-approximate PCP and PCR. Therefore, our pseudocodes allow users to set $\gamma = 0$ and thus ignore this parameter $\gamma$; in such a case, we shall use $\gamma = \log(n)/n$ which is equivalent to setting $\gamma = \Theta(\varepsilon)$ because $n = \Theta(\gamma^{-1} \log(1/\gamma\varepsilon))$.

## 8 Experiments

In the same way as [13], we conclude this paper with an empirical evaluation to demonstrate our theorems.

**Datasets.** We consider synthetic and real-life datasets.

- We generate the synthetic dataset in the same way as [13]. That is, we form a $3000 \times 2000$ dimensional matrix $\mathbf{A}$ via the SVD $\mathbf{A} = \mathbf{U}\mathbf{\Sigma}\mathbf{V}^\top$ where $\mathbf{U}$ and $\mathbf{V}$ are random orthonormal matrices and $\mathbf{\Sigma}$ contains random singular values. Among the 2000 singular values, we let half of them be randomly chosen from $[0, \sqrt{0.1}(1-a)]$ and the other half randomly chosen from $[\sqrt{0.1}(1+a), 1]$. We generate vector $b$ by adding noise to the response $\mathbf{A}x$ of a random "true" x that correlates with $\mathbf{A}$'s top principal components. We consider eigenvalue threshold $\lambda = 0.1$, and use $a = 0, 0.01, 0.02, 0.1$ in our experiments. We call these datasets `random-`$a$.

- As for the real-life dataset, we use `mnist` [11]. After scaling its largest singular value to one,[9] we choose the eigenvalue threshold $\lambda = 0.0025$ (or equivalently singular value threshold $\sqrt{\lambda} = 0.05$). The closest singular values to this threshold are respectively 0.05027 and 0.04958.

**Algorithms.** We implemented our algorithm and Frostig *et al.* [13] (which we call FMMS for short) and minimized the number of calls to ridge regression in our implementations. For instance, if using our pseudocode `QuickPCP`, the number of ridge regression calls is $2n+1$; if using our pseudocode `QuickPCR`, the number of extra ridge regression calls is $m+1$. We choose $m = 10$ in all of our experiments because the theoretical prediction of $m$ is only a small logarithmic quantity (see Lemma 3.4 and Lemma 3.5).

We also implemented a practical heuristic using Krylov subspace that were found on the website [12]. We call this algorithm Krylov method for short. Krylov method transforms the covariance matrix $\mathbf{A}\mathbf{A}^\top$ into a lower-dimensional Krylov subspace and performs exact PCP and PCR there. Similar to this paper, Krylov method also reduces PCP and PCR to multiple calls of ridge regressions.[10]

We emphasize that Krylov method has *no supporting theory* behind it. Since we find it performs much faster than FMMS in practice, we include it in our experiments for a stronger comparison.

*Remark* 8.1. There are two main issues behind the missing theory of Krylov method.

- Stability. If matrix-vector multiplications are only approximate, Krylov-based methods are usually unstable so one needs to replace it with other stable variants. Our polynomial approximation $g_n(x)$ can be viewed as one such stable variant.
- Accuracy. To the best of our knowledge, even with exact computations, if there is no eigengap around threshold $\lambda$ —which is usually the case in real life— it is unlikely that Krylov method can achieve a $\log(1/\varepsilon)$ convergence with respect to the $\varepsilon$-parameter in $(\gamma,\varepsilon)$-approximate PCP or PCR.[11] Our experiments later (namely Figure 3(c) and 3(f)) shall also confirm on this.

## 8.1 Evaluation 1: With Eigengap Assumption

In the first evaluation we consider matrices that satisfy the eigengap assumption. To simulate an eigengap, we use random datasets `random-`$a$ with $a = 0.01, 0.02, 0.1$ and present our findings in Figure 2 in terms of the following three performance measures:

- Regression Error: $\|x - x^*\|_2/\|x^*\|_2$; where $x$ is the output of a PCR algorithm and $x^* = (\mathbf{A}^\top\mathbf{A})^\dagger\mathbf{P}_\lambda\mathbf{A}^\top b$ is the exact PCR solution.
- Projection Error: $\|\xi - \xi^*\|_2/\|\xi^*\|_2$; where $\xi$ is the output of a PCP algorithm and $\xi^* = \mathbf{P}_\lambda\mathbf{A}^\top b$ is the exact PCP solution.
- Denoising Error: $\|(\mathbf{I} - \mathbf{P}_\lambda)\xi\|_2/\|\xi\|_2$; where $\xi$ is the output of a PCP algorithm.

The $x$-axis of these plots represent the number of calls to ridge regression, and in Figure 2 we use exact implementations of ridge regression similar to the experiments in [13]. Note that the horizontal axis starts with 0 for projection performances (second and third column) and with 10 for regression performance (first column). This is so because in order to reduce PCR to PCP one needs $m+1$ calls to ridge regression in `QuickPCR` and in our experiments we simply choose $m = 10$.

[9] This is a cheap procedure and for instance can be done by power method [13].

[10] The original code [12], when working with Krylov subspace of dimension $k$, requires $2k$ calls of ridge regression. In our experiments, we improved this implementation and reduced it from $2k$ calls to $k$ calls for a stronger comparison.

[11] This is so because Krylov method works in a smaller dimension whose so-called "Ritz values" approximate the original eigenvalues of $\mathbf{A}^\top\mathbf{A}$. However, this approximation cannot be "exponentially close" because there are only very few Ritz values as compared to the original eigenvalues of $\mathbf{A}^\top\mathbf{A}$.

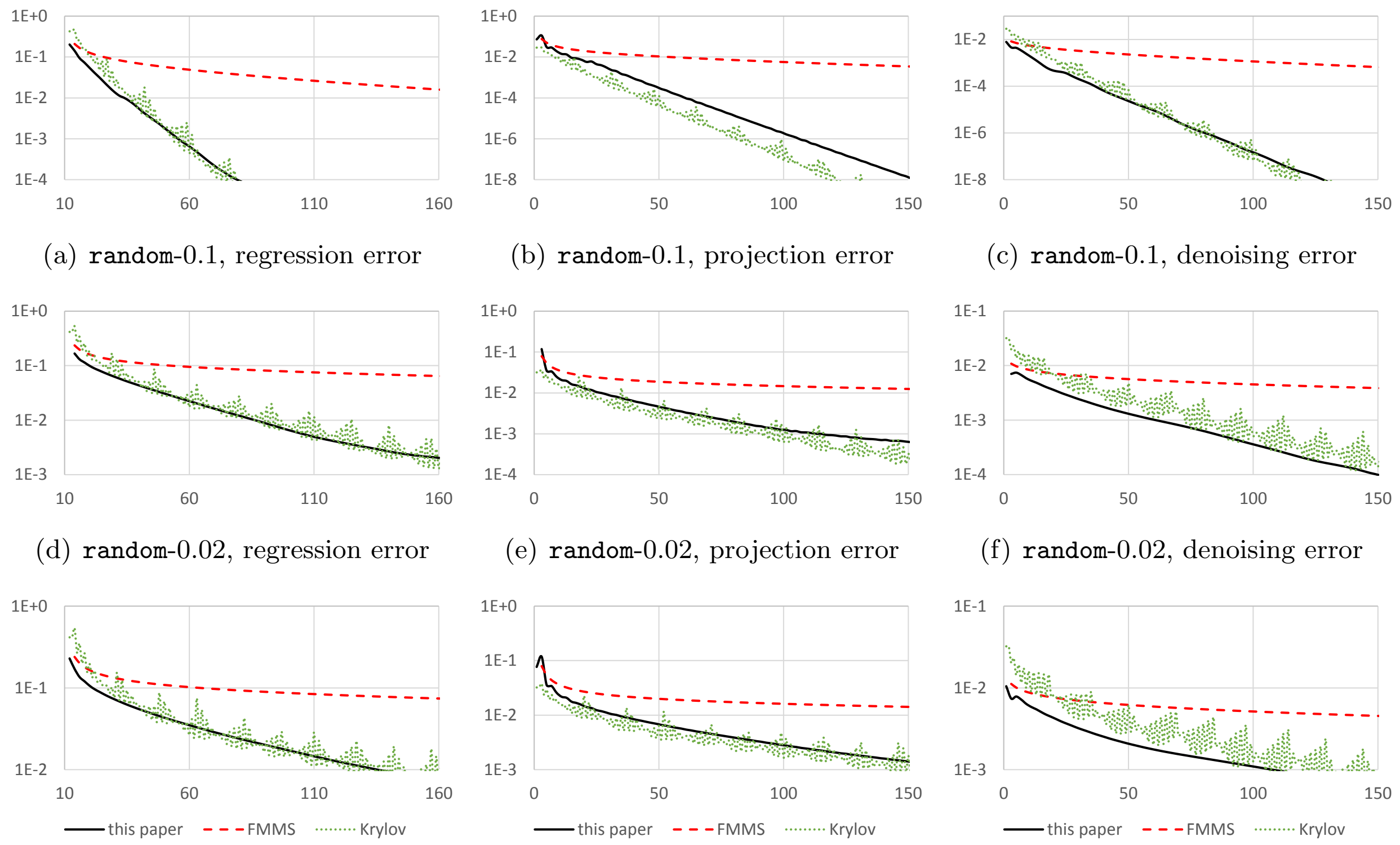

(a) `random`-0.1, regression error (b) `random`-0.1, projection error (c) `random`-0.1, denoising error

(d) `random`-0.02, regression error (e) `random`-0.02, projection error (f) `random`-0.02, denoising error

(g) `random`-0.01, regression error (h) `random`-0.01, projection error (i) `random`-0.01, denoising error

Figure 2: Performance comparison on `random`-$a$ datasets with eigengap $a > 0$.
*In the plots, the x-axis represents the number of oracle calls to ridge regression and the y-axis represents performance. Denoting by $x$ and $\xi$ respectively the PCR and PCP outputs, then regression error is $\|x - x^*\|_2/\|x^*\|_2$, projection error is $\|\xi - \xi^*\|_2/\|\xi^*\|_2$, and denoising error is $\|(\mathbf{I} - \mathbf{P}_\lambda)\xi\|_2/\|\xi\|_2$.*

We make some important observations from these results

- We significantly outperform FMMS for our choices of $a$.
- Our performance degrades as $a$ (and thus $\gamma$) decreases; this is consistent to our theory.
- The performance of Krylov method fluctuates partly due to the missing theory behind it. This limits the practicality of Krylov method, because it is hardly possible for the algorithm to determine when is the best time to stop the algorithm.[12]
- If the fluctuation of Krylov method is ignored, it matches the performance of `QuickPCP` and `QuickPCR`. This is an interesting phenomenon and might even be a first evidence towards a theoretical proof for Krylov method.

## 8.2 Evaluation 2: Without Eigengap Assumption

In our second evaluation we consider scenarios when there is no significant eigengap around the projection threshold $\lambda$. We consider dataset `random`-$a$ for $a = 0$ as well as dataset `mnist`. This

[12] Of course, if the true projection matrix $\mathbf{P}_\lambda$ is given explicitly, we can determine a good iteration to stop. However, the entire PCP problem is regarding how to compute $\mathbf{P}_\lambda$ without explicitly constructing it.

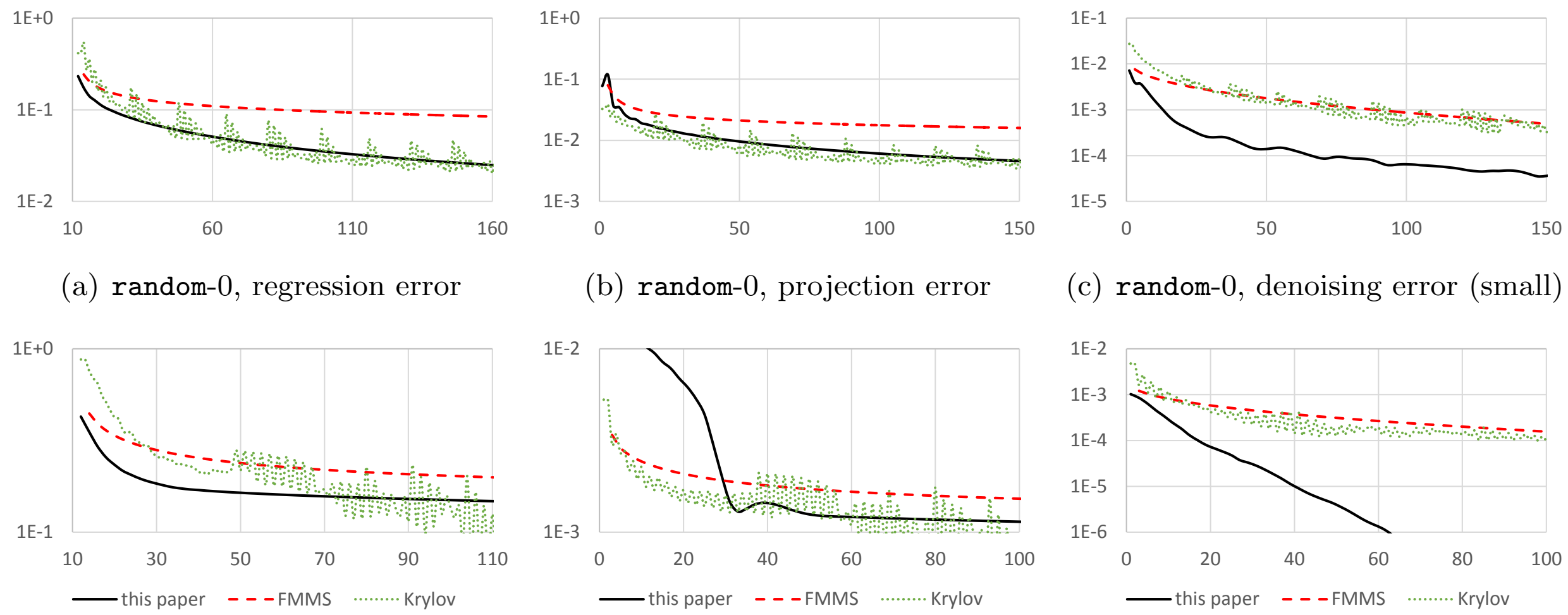

(a) **`random`**-0, regression error (b) **`random`**-0, projection error (c) **`random`**-0, denoising error (small)

(d) **`mnist`**, regression error (e) **`mnist`**, projection error (f) **`mnist`**, denoising error (small)

Figure 3: Gap-free performance comparison on **`random`**-0 and **`mnist`**.
*In the plots, the $x$-axis represents the number of oracle calls to ridge regression and the $y$-axis represents performance. Denoting by $x$ and $\xi$ respectively the PCR and PCP outputs, then regression error is $\|x - x^*\|_2/\|x^*\|_2$, projection error is $\|\xi - \xi^*\|_2/\|\xi^*\|_2$, and denoising error (small) is $\|(\mathbf{I} - \mathbf{P}_{0.81\lambda})\xi\|_2/\|\xi\|_2$.*

time, we also consider three performance measures. The first two are the same as the previous subsection, as for the third measure, we replace it with

- DENOISING ERROR (SMALL): $\|(\mathbf{I} - \mathbf{P}_{0.81\lambda})\xi\|_2/\|\xi\|_2$.

We emphasize here that in gap-free scenarios, regression error, projection error, or even the quantity $\|(\mathbf{I} - \mathbf{P}_\lambda)\xi\|_2$ can all be very large — in the extreme case if there is an eigenvector that has exactly eigenvalue $\lambda$, then these quantities do not converge to zero. This is why our gap-free approximation definitions do not account for such quantities (see Def. 3.1 and Def. 3.2).

In contrast, by focusing only on eigenvectors that are less than threshold $(1-\gamma)\lambda$ for some $\gamma > 0$, and looking at $\|(\mathbf{I} - \mathbf{P}_{(1-\gamma)\lambda})\xi\|_2$, this quantity can *indeed converge* to $\varepsilon > 0$ with a speed that is $O(\gamma^{-1}\log(1/\varepsilon))$ if our algorithm is used (see Theorem 7.4). Note that this speed was only $O(\gamma^{-2}\log(1/\varepsilon))$ for FMMS.

We present our findings in Figure 3 and make some important conclusions here:

- Our method still significantly outperforms FMMS.
- In terms of denoising error, our method significantly outperforms Krylov method. This is so because, according to Remark 8.1, Krylov method cannot achieve a $\log(1/\varepsilon)$ convergence rate with respect to the $\varepsilon$-parameter in $(\gamma, \varepsilon)$-approximate PCP or PCR. Threfore, our method is clearly the best for denoising purposes.

## 8.3 Evaluation 3: Stability Test

In our third evaluation, we verify that our method continues to work well even if ridge regressions are computed with moderate error. We consider two types of errors in our experiments:

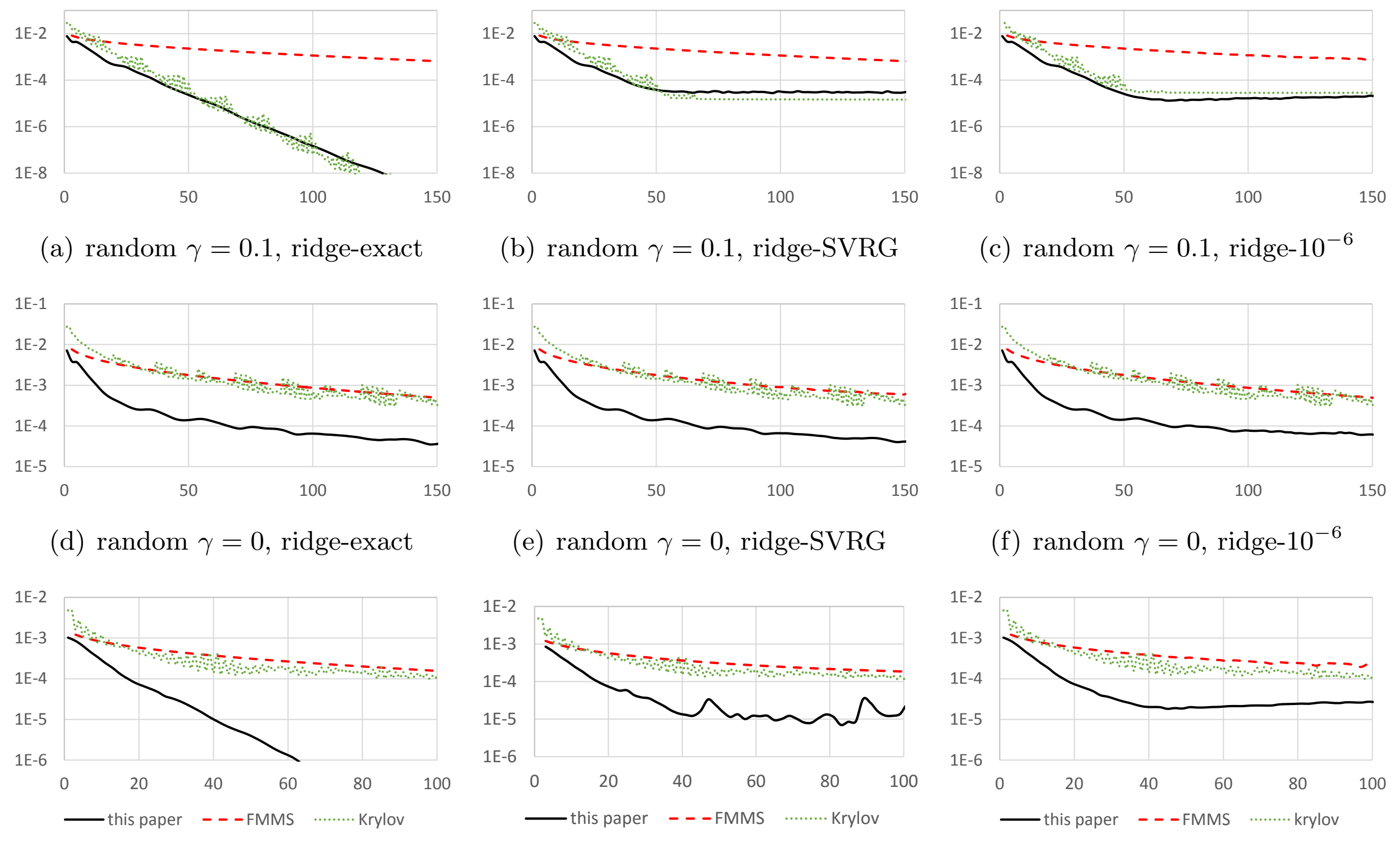

(a) random $\gamma = 0.1$, ridge-exact (b) random $\gamma = 0.1$, ridge-SVRG (c) random $\gamma = 0.1$, ridge-$10^{-6}$

(d) random $\gamma = 0$, ridge-exact (e) random $\gamma = 0$, ridge-SVRG (f) random $\gamma = 0$, ridge-$10^{-6}$

(g) mnist, ridge-exact (h) mnist, ridge-SVRG (i) mnist, ridge-$10^{-7}$

Figure 4: Stability test — exact vs. approximate ridge regression subroutines.
*In the plots, the x-axis represents the number of oracle calls to ridge regression and the y-axis represents the denoising error. We compare exact implementation of ridge regression with ridge-SVRG and ridge-*$10^{-k}$.

*Remark.* Although it *seems* our method is more affected by error than FMMS, we emphasize that this is because FMMS is *too slow* and still works in a very low-accuracy regime in the plots. (For instance, as a stable algorithm, FMMS should not be affected by error of magnitude around $10^{-6}$ when the desired accuracy is above $10^{-4}$.)

- ridge-SVRG: we run the SVRG [17] method for 50 passes to solve each ridge regression.[13]
- ridge-$10^{-k}$: we run exact ridge regression but randomly add noise $[-10^{-k}, 10^{-k}]$ per coordinate.

We present our findings in Figure 4. For cleanness, we compare only the denoising error and only on datasets `mnist`, `random`-0 and `random`-0.1.[14] We make the following conclusions and remarks:

- Even with inexact ridge regression, our method still works very well. We continue to outperform FMMS significantly.
- Compared with Krylov method, we continue to outperform it significantly in gap-free scenarios.
- Although it *seems* our method is more affected by error than FMMS, we emphasize that this is because FMMS is *too slow* and still works in a very low-accuracy regime in the plots. (For instance, as a stable algorithm, FMMS should not be affected by error of magnitude around $10^{-6}$ when the desired accuracy is above $10^{-4}$.)

[13] We choose the epoch length of SVRG to be $2n$, and therefore full gradients are computed every $2n$ stochastic iterations. Each $n$ stochastic iterations is counted as one "pass" of the data, and each full gradient computation is counted as one "pass" of the data.

[14] Since `mnist` and `random`-0 are datasets without significant eigengap, we present "denoising error (small)" as defined in Section 8.2.

## 9 Conclusion

We summarize our contributions.

- We put forward approximate notions for PCP and PCR that do not rely on any eigengap assumption. Our notions reduce to standard ones under the eigengap assumption.
- We design near-optimal polynomial approximation $g(x)$ to $\mathrm{sgn}(x)$ satisfying (1.1) and (1.2).
- We develop general stable recurrence formula for matrix Chebyshev polynomials; as a corollary, our $g(x)$ can be applied to matrices in a stable manner.
- We obtain faster, provable PCA-free algorithms for PCP and PCR than known results.

## Acknowledgements

We thank Yin Tat Lee for suggesting us the new title, and anonymous referees for useful suggestions. Z. Allen-Zhu is partially supported an NSF Grant, no. CCF-1412958, and a Microsoft Research Grant, no. 0518584. Any opinions, findings and conclusions or recommendations expressed in this material are those of the author(s) and do not necessarily reflect the views of NSF or Microsoft.

# Appendix

## A Proof of Lemma 3.5

**Lemma 3.5.** *For fixed $\lambda, \varepsilon \in (0,1)$ and $\gamma \in (0, 2/3]$, let $\mathbf{A}$ be a matrix whose singular values are no more than 1. Let* `ApxRidge` *be any $O(\frac{\varepsilon}{m^2})$-approximate ridge regression solver, and $\mathcal{B}$ be any $(\gamma, O(\frac{\varepsilon\lambda}{m^2}))$-approximate PCP solver. Then, procedure (3.1) satisfies,*

$$\left\{\begin{array}{c} \|(\mathbf{I} - \mathbf{P}_{(1-\gamma)\lambda}) s_m\| \le \varepsilon\|b\| \quad \text{, and} \\ \|\mathbf{A} s_m - b\| \le \|\mathbf{A}(\mathbf{A}^\top\mathbf{A})^\dagger \mathbf{P}_{(1+\gamma)\lambda}\mathbf{A}^\top b - b\| + \varepsilon\|b\| \end{array}\right\} \quad \text{if} \quad m = \Theta(\log(1/\varepsilon\gamma))$$

*Proof of Lemma 3.5.* We first notice that the approximation guarantee of $\mathcal{B}$ implies

$$\|s_0\| = \|\mathcal{B}(\mathbf{A}^\top b)\| \le \|\mathbf{A}^\top b\| + O\left(\varepsilon\lambda/m^2\right)\|b\| \le 2\|b\| \ .$$

Let us consider a new exact sequence $\{s_k^*\}_{k\ge 0}$ where

$$s_0^* = \mathbf{P}_{(1-\gamma)\lambda} s_0, \quad s_1^* = (\mathbf{A}^\top\mathbf{A} + \lambda\mathbf{I})^{-1} s_0^*, \quad \forall k \ge 1\colon s_{k+1}^* = s_1^* + \lambda\cdot(\mathbf{A}^\top\mathbf{A} + \lambda\mathbf{I})^{-1} s_k^* \ .$$

**Step I.** We first bound the error between $s_k$ and $s_k^*$. We have $\|s_{k+1}^*\| \le \|s_1^*\| + \lambda\|\mathbf{A}^\top\mathbf{A} + \lambda\mathbf{I}\|_2\|s_k^*\| \le \|s_1^*\| + \|s_k^*\|$ which implies $\|s_k^*\| \le k\|s_1^*\| \le \frac{k}{\lambda}\|s_0^*\| \le \frac{2k}{\lambda}\|b\|$. Therefore,

$$\begin{aligned} \|s_{k+1}^* - s_{k+1}\| &\le \|s_1^* - s_1\| + \lambda\|(\mathbf{A}^\top\mathbf{A} + \lambda\mathbf{I})^{-1} s_k^* - \texttt{ApxRidge}(\mathcal{A}, \lambda, s_k)\| \\ &\le \|s_1^* - s_1\| + \lambda\|(\mathbf{A}^\top\mathbf{A} + \lambda\mathbf{I})^{-1}(s_k^* - s_k)\| + \lambda\|(\mathbf{A}^\top\mathbf{A} + \lambda\mathbf{I})^{-1} s_k - \texttt{ApxRidge}(\mathcal{A}, \lambda, s_k)\| \\ &\le \|s_1^* - s_1\| + \|s_k^* - s_k\| + O\left(\lambda\varepsilon/m^2\right)\|s_k\| \ . \end{aligned} \tag{A.1}$$

Since $\|s_k^*\| \le \frac{2k}{\lambda}\|b\| \le \frac{2m}{\lambda}\|b\|$ and since $\|s_0^* - s_0\| \le O(\frac{\varepsilon\lambda}{m^2})\|b\|$, we can conclude from (A.1) (by telescoping sum over $k = 1, \dots, k'$) that

$$\forall k' \in \{0, 1, \dots, m\}\colon \quad \|s_{k'}^* - s_{k'}\| \le \varepsilon\|b\| \ .$$

**Step II.** We next focus on $s_k^*$ and decompose $s_k^*$ into three parts: for every $k \geq 0$, define

$$v_{1,k} = \mathbf{P}_{(1+\gamma)\lambda} s_k^* =: \mathbf{P}_1 s_k^*, \quad v_{2,k} = (\mathbf{I} - \mathbf{P}_{(1-\gamma)\lambda}) s_k^* =: \mathbf{P}_2 s_k^*, \quad v_{3,k} = (\mathbf{P}_{(1-\gamma)\lambda} - \mathbf{P}_{(1+\gamma)\lambda}) s_k^* =: \mathbf{P}_3 s_k^* \ .$$

The update rule of $s_k^*$ tells us that

$$\forall i \in [3], k \geq 1: \quad v_{i,k} = \frac{1}{\lambda} \sum_{t=1}^{k} \big(\lambda(\mathbf{A}^\top \mathbf{A} + \lambda \mathbf{I})^{-1} \mathbf{P}_i\big)^t s_0^* \ .$$

In particular, since $v_{2,0} = \mathbf{P}_2 s_0^* = 0$ we always have $v_{2,m} = 0$.

As for $v_{1,m}$ and $v_{3,m}$, we first notice that if we denote by $p_k(x) \stackrel{\text{def}}{=} \sum_{t=1}^k \lambda^{t-1} x^t$, then $v_{i,k} = p_k((\mathbf{A}^\top \mathbf{A} + \lambda \mathbf{I})^{-1}) \mathbf{P}_i s_0^*$. But since $\lim_{k\to\infty} p_k(x) = \frac{x}{1-\lambda x} =: p(x)$, we have

$$\lim_{k\to\infty} v_{i,k} = p((\mathbf{A}^\top \mathbf{A} + \lambda \mathbf{I})^{-1}) \mathbf{P}_i s_0^* = (\mathbf{A}^\top \mathbf{A})^\dagger \mathbf{P}_i s_0^* = (\mathbf{A}^\top \mathbf{A})^\dagger v_{i,0} \ .$$

At the same time, note that the spectral norms $\|\lambda \cdot (\mathbf{A}^\top \mathbf{A} + \lambda \mathbf{I})^{-1} \mathbf{P}_1\|_2$ and $\|\lambda \cdot (\mathbf{A}^\top \mathbf{A} + \lambda \mathbf{I})^{-1} \mathbf{P}_3\|_2$ are both no more than $\frac{3}{4}$. (This is so because for every eigenvalue $\lambda_j$ of $\mathbf{A}^\top \mathbf{A}$ that is below $\lambda(1-\gamma)$ we have $\frac{\lambda}{\lambda+\lambda_j} \leq \frac{\lambda}{\lambda + (1/3)\lambda} = \frac{3}{4}$.) Therefore, for both $i = 1$ and $i = 3$, we have

$$\Big\| v_{i,m} - \lim_{k\to\infty} v_{i,k} \Big\| \leq \frac{1}{\lambda} \sum_{t=m+1}^{\infty} \|\lambda \cdot (\mathbf{A}^\top \mathbf{A} + \lambda \mathbf{I})^{-1} \mathbf{P}_i\|_2^t \cdot \|s_0^*\| \leq (3/4)^m \cdot O\big(\|b\|/\lambda\big) \ .$$

In other words, choosing $m = \Theta(\log(1/\varepsilon\lambda))$, we have

$$\|v_{1,m} - (\mathbf{A}^\top \mathbf{A})^\dagger v_{1,0}\| \leq \varepsilon \|b\| \quad \text{and} \quad \|v_{3,m} - (\mathbf{A}^\top \mathbf{A})^\dagger v_{3,0}\| \leq \varepsilon \|b\| \ . \tag{A.2}$$

**Step III.** We now take into account the error of the PCP solver $\mathcal{B}$. For $v_{1,m}$, we have:

$$\begin{aligned} \|v_{1,m} - (\mathbf{A}^\top \mathbf{A})^\dagger \mathbf{P}_1 \mathbf{A}^\top b\| &\leq \|v_{1,m} - (\mathbf{A}^\top \mathbf{A})^\dagger v_{1,0}\| + \|(\mathbf{A}^\top \mathbf{A})^\dagger \mathbf{P}_1 (\mathcal{B}(\mathbf{A}^\top b) - \mathbf{A}^\top b)\| \\ &\leq \varepsilon \|b\| + \frac{1}{\lambda} \|\mathbf{P}_1 \big(\mathcal{B}(\mathbf{A}^\top b) - \mathbf{A}^\top b\big)\| \leq 2\varepsilon \|b\| \ , \end{aligned} \tag{A.3}$$

where the first inequality uses triangle inequality, the second uses (A.2), and the third uses Def. 3.1 and $\|\mathbf{A}^\top b\| \leq \|b\|$.

As for $v_{3,m}$, we let $\mathbf{A} = \mathbf{U} \mathbf{\Sigma} \mathbf{V}^\top$ be the SVD of $\mathbf{A}$ and let $\mathbf{\Sigma}^\dagger$ be the same matrix $\mathbf{\Sigma}$ except all non-zero elements get inverted. We have

$$\begin{aligned} \|\mathbf{A}(v_{3,m} - (\mathbf{A}^\top \mathbf{A})^\dagger \mathbf{P}_3 \mathbf{A}^\top b)\| &\overset{①}{\leq} \Big\|\mathbf{A}\big((\mathbf{A}^\top \mathbf{A})^\dagger v_{3,0} - (\mathbf{A}^\top \mathbf{A})^\dagger \mathbf{P}_3 \mathbf{A}^\top b\big)\Big\| + \|\mathbf{A} v_{3,m} - \mathbf{A}(\mathbf{A}^\top \mathbf{A})^\dagger v_{3,0}\| \\ &\overset{②}{\leq} \Big\|\mathbf{A}\big((\mathbf{A}^\top \mathbf{A})^\dagger v_{3,0} - (\mathbf{A}^\top \mathbf{A})^\dagger \mathbf{P}_3 \mathbf{A}^\top b\big)\Big\| + \varepsilon \|b\| \\ &= \Big\|\mathbf{U} \mathbf{\Sigma}^\dagger \mathbf{V}^\top \mathbf{P}_3 \big(\mathcal{B}(\mathbf{A}^\top b) - \mathbf{A}^\top b\big)\Big\| + \varepsilon \|b\| \\ &\overset{③}{\leq} \Big\|\mathbf{\Sigma}^\dagger \mathbf{V}^\top \mathbf{P}_3 \Big(\mathcal{B}(\mathbf{A}^\top b) - \mathbf{A}^\top b\Big)\Big\| + \varepsilon \|b\| \\ &= \sum_{i: \lambda_i \in [(1-\gamma)\lambda, (1+\gamma)\lambda]} \frac{1}{\sqrt{\lambda_i}} |\langle v_i, \mathcal{B}(\mathbf{A}^\top b) - \mathbf{A}^\top b \rangle| + \varepsilon \|b\| \\ &\overset{④}{\leq} \sum_{i: \lambda_i \in [(1-\gamma)\lambda, (1+\gamma)\lambda]} \frac{1}{\sqrt{\lambda_i}} |\langle v_i, \mathbf{A}^\top b \rangle| + 2\varepsilon \|b\| \\ &= \|(\mathbf{A}^\top \mathbf{A})^\dagger \mathbf{P}_3 \mathbf{A}^\top b\| + 2\varepsilon \|b\| \ . \end{aligned} \tag{A.4}$$

Above, ① uses triangle inequality, ② uses (A.2) and the fact $\|\mathbf{A}\|_2 \leq 1$, ③ uses $\|\mathbf{U}\|_2 \leq 1$, ④ uses Def. 3.1 and $\|\mathbf{A}^\top b\| \leq \|b\|$.

**Step IV.** Finally we put everything together and bound the regression error. Denote by $\mathsf{opt} = \|\mathbf{A}(\mathbf{A}^\top\mathbf{A})^\dagger\mathbf{P}_{(1+\gamma)\lambda}\mathbf{A}^\top b - b\|$. If we decompose $b$ as

$$b = \Big(\sum_{i=1}^{3}\mathbf{A}(\mathbf{A}^\top\mathbf{A})^\dagger\mathbf{P}_i\mathbf{A}^\top b\Big) + (b - \mathbf{A}(\mathbf{A}^\top\mathbf{A})^\dagger\mathbf{A}^\top b) \ , \tag{A.5}$$

then the four vectors in (A.5) are orthogonal to each other, which gives us

$$\begin{aligned} \mathsf{opt} &= \|\mathbf{A}(\mathbf{A}^\top\mathbf{A})^\dagger\mathbf{P}_1\mathbf{A}^\top b - b\| \\ &= \|\mathbf{A}(\mathbf{A}^\top\mathbf{A})^\dagger\mathbf{P}_2\mathbf{A}^\top b\| + \|\mathbf{A}(\mathbf{A}^\top\mathbf{A})^\dagger\mathbf{P}_3\mathbf{A}^\top b\| + \|\mathbf{A}(\mathbf{A}^\top\mathbf{A})^\dagger\mathbf{A}^\top b - b\| \ . \end{aligned} \tag{A.6}$$

Now we compute the regression error with respect to $s_m^*$:

$$\begin{aligned} \|\mathbf{A}s_m^* - b\| &\overset{①}{=} \|\mathbf{A}(v_{1,m} + v_{3,m}) - b\| \\ &\overset{②}{=} \Big\|\mathbf{A}(v_{1,m} + v_{3,m}) - \sum_{i=1}^{3}\mathbf{A}(\mathbf{A}^\top\mathbf{A})^\dagger\mathbf{P}_i\mathbf{A}^\top b + (b - \mathbf{A}(\mathbf{A}^\top\mathbf{A})^\dagger\mathbf{A}^\top b)\Big\| \\ &\overset{③}{\le} \Big\|\mathbf{A}(v_{1,m} - (\mathbf{A}^\top\mathbf{A})^\dagger\mathbf{P}_1\mathbf{A}^\top b)\Big\| + \Big\|\mathbf{A}(v_{3,m} - (\mathbf{A}^\top\mathbf{A})^\dagger\mathbf{P}_3\mathbf{A}^\top b)\Big\| \\ &\quad + \|\mathbf{A}(\mathbf{A}^\top\mathbf{A})^\dagger\mathbf{P}_2\mathbf{A}^\top b\| + \|\mathbf{A}(\mathbf{A}^\top\mathbf{A})^\dagger\mathbf{A}^\top b - b\| \\ &\overset{④}{\le} 4\varepsilon\|b\| + \|\mathbf{A}(\mathbf{A}^\top\mathbf{A})^\dagger\mathbf{P}_2\mathbf{A}^\top b\| + \|\mathbf{A}(\mathbf{A}^\top\mathbf{A})^\dagger\mathbf{P}_3\mathbf{A}^\top b\| + \|\mathbf{A}(\mathbf{A}^\top\mathbf{A})^\dagger\mathbf{A}^\top b - b\| \\ &\overset{⑤}{=} \mathsf{opt} + 4\varepsilon\|b\| \ . \end{aligned}$$

Above, ① is because $v_{2,m} = 0$; ② uses (A.5); ③ uses triangle inequality; ④ uses (A.3) and (A.4); ⑤ uses (A.6).

Finally, using $\|s_m^* - s_m\| \le \varepsilon\|b\|$ we complete the proof that $\|\mathbf{A}s_m - b\| \le \mathsf{opt} + 5\varepsilon\|b\|$. We also have $\|\mathbf{P}_2 s_m\| \le \varepsilon\|b\| + \|\mathbf{P}_2 s_m^*\| = \varepsilon\|b\|$ because $\mathbf{P}_2 s_m^* = v_{2,m} = 0$. □

# B Appendix for Section 4

**Lemma 4.1.** *Suppose $f(z)$ is analytic on $\mathcal{E}_\rho$ and for every $k \ge 0$, $f^{(k)}(0) \ge 0$. Then, for every $n \in \mathbb{N}$, letting $p_n(x)$ and $q_n(x)$ be be the degree-$n$ Chebyshev truncated series and Chebyshev interpolation of $f(x)$, we have*

$$\forall y \in [0, \rho]: \quad 0 \le p_n(1+y), q_n(1+y) \le f(1+y) \ .$$

To show Lemma 4.1 we first need an auxiliary lemma, which can be proved by some careful case analysis (see Appendix B.1).

**Lemma B.1.** *Let $m, n \in \mathbb{N}$ be two integers, then $a_{m,n} = \int_{-1}^{1}\frac{x^m}{\sqrt{1-x^2}}\mathcal{T}_n(x)dx \ge 0$.*

Lemma B.1 essentially says that the Chebyshev coefficients of any function $x^m$ must be all non-negative. We also recall the following lemma regarding high-order derivatives of Chebyshev truncated series:

**Lemma B.2** (cf. Theorem 21.1 of [23])**.** *Suppose $f(z)$ is analytic on $\mathcal{E}_\rho$ with $\rho > 0$, and let $p_n(x)$ be the degree-$n$ Chebyshev truncated series of $f(x)$. Then, for every $k \ge 0$,*

$$\lim_{n\to+\infty}\max_{x\in[-1,1]}\Big\{|f^{(k)}(x) - p_n^{(k)}(x)|\Big\} = 0 \ .$$

We are now ready to prove Lemma 4.1. The main idea is to expand $f$ into its Taylor series, and then deal with monomials $x^m$ one by one:

*Proof of Lemma 4.1.* Since $f^{(k)}(0) \geq 0$ for all $k \geq 0$, and since $f(z)$ is analytic, we can write $f$ as $f(z) = \sum_{k=0}^{\infty} r_k z^k$ where each $r_k$ is a nonnegative real. Consider the $i$-th coefficient of Chebyshev series:

$$a_i = \frac{2 - \mathbb{1}[i=0]}{\pi} \int_{-1}^{1} \frac{f(x)}{\sqrt{1-x^2}} \mathcal{T}_i(x) dx = \frac{2 - \mathbb{1}[i=0]}{\pi} \sum_{k=0}^{\infty} r_k \int_{-1}^{1} \frac{x^k}{\sqrt{1-x^2}} \mathcal{T}_i(x) \geq 0$$

where the last inequality is due to Lemma B.1, and the integral and infinite Taylor sum are interchangeable.[15] This implies we can write $p_n(x) = \sum_{i=0}^{n} a_i \mathcal{T}_i(x)$ where each $a_i \geq 0$.

Since each $\mathcal{T}_i(1+y)$ is a polynomial of degree $i$, it exactly equals to its degree-$i$ Taylor expansion $\sum_{k=0}^{i} \frac{y^k}{k!} \mathcal{T}_i^{(k)}(1)$. Thus, we have (recall $y \in [0, \rho]$)

$$p_n(1+y) = \sum_{i=0}^{n} a_i \mathcal{T}_i(1+y) = \sum_{i=0}^{n} \sum_{k=0}^{i} \frac{a_i}{k!} \mathcal{T}_i^{(k)}(1) y^k = \sum_{k=0}^{n} \frac{1}{k!} \left( \sum_{i=k}^{n} a_i \mathcal{T}_i^{(k)}(1) \right) y^k \ .$$

Denote by $b_{k,n} = \left( \sum_{i=k}^{n} a_i \mathcal{T}_i^{(k)}(1) \right)$. Since for every $i, k \geq 0$ it satisfies $T_i^{(k)}(1) \geq 0$ (which is a factual property of Chebyshev polynomial) and $a_i \geq 0$, we know $b_{k,n} \geq 0$ and moreover $b_{k,n}$ is monotonically non-decreasing in $n$ for each $k \geq 0$. On the other hand, Lemma B.2 implies

$$\lim_{n \to \infty} \left| p_n^{(k)}(1) - f^{(k)}(1) \right| = \lim_{n \to \infty} \left| b_{k,n} - f^{(k)}(1) \right| = 0 \ ,$$

so we must have $0 \leq b_{k,n} \leq f^{(k)}(1)$ for every $n \in \mathbb{N}$ (because $b_{k,n}$ is non-decreasing in $n$). Therefore, for every $y \in [0, \rho]$:

$$0 \leq p_n(1+y) = \sum_{k=0}^{n} \frac{1}{k!} b_{k,n} y^k \leq \sum_{k=0}^{\infty} \frac{1}{k!} b_{k,n} y^k \leq \sum_{k=0}^{\infty} \frac{1}{k!} f^{(k)}(1) y^k = f(1+y) \ . \tag{B.1}$$

Finally, since $q_n(x) \stackrel{\text{def}}{=} \sum_{k=0}^{n} c_k \mathcal{T}_k(x)$ is a degree-$n$ Chebyshev interpolation polynomial, the aliasing Lemma 2.8 tells us $c_i \geq 0$ for every $i = 0, 1, \ldots, n$. Furthermore, applying the aliasing Lemma 2.8 again we have $c_i \geq a_i$ for $i = 0, 1, \ldots, n$ but $\sum_{i=0}^{n} c_i = \sum_{i=0}^{\infty} a_i$. Therefore, using the fact that $\mathcal{T}_i^{(k)}(1)$ is a monotone increasing function in $i$ (for every fixed $k$), we have

$$0 \leq \sum_{i=0}^{n} c_i \mathcal{T}_i^{(k)}(1) \leq \sum_{i=0}^{\infty} a_i \mathcal{T}_i^{(k)}(1) = \lim_{n \to \infty} b_{k,n} = f^{(k)}(1) \ .$$

Finally, an analogous proof as (B.1) also shows $0 \leq q_n(1+y) \leq f(1+y)$ for every $y \in [0, \rho]$. □

## B.1 Proof of Lemma B.1

**Lemma B.1.** *Let $m, n \in \mathbb{N}$ be two integers, then $a_{m,n} = \int_{-1}^{1} \frac{x^m}{\sqrt{1-x^2}} \mathcal{T}_n(x) dx \geq 0$.*

[15]The interchangeability and be verified as follows. Denoting by $f_m(x) \stackrel{\text{def}}{=} \sum_{k=0}^{m} r_m x^m$, we have $f_m(x)$ uniformly converges to $f(x)$ on $x \in [-1, 1]$ because the Taylor expansion of any analytical function has local uniform convergence, but $[-1, 1]$ is a compact, closed interval so local uniform convergence becomes global uniform convergence.

For every $\varepsilon > 0$, let $M$ be the integer so that for every $m \geq M$ it satisfies $\max_{x \in [-1,1]} |f_m(x) - f(x)| \leq \varepsilon$. We compute that $\left| \int_{-1}^{1} \frac{f(x)}{\sqrt{1-x^2}} \mathcal{T}_i(x) dx - \sum_{k=0}^{m} r_k \int_{-1}^{1} \frac{x^k}{\sqrt{1-x^2}} \mathcal{T}_i(x) dx \right| = \left| \int_{-1}^{1} \frac{f(x) - f_m(x)}{\sqrt{1-x^2}} \mathcal{T}_i(x) dx \right| \leq \int_{-1}^{1} \frac{\varepsilon}{\sqrt{1-x^2}} dx = \varepsilon\pi$. Therefore, the left hand side converges to zero so the integral and the infinite Taylor sum are interchangeable.

*Proof of Lemma B.1.* Recall that $\mathcal{T}_n(-x) = (-1)^n \mathcal{T}_n(x)$. Therefore,

$$a_{m,n} = \int_1^{-1} \frac{(-x)^m}{\sqrt{1-x^2}} \mathcal{T}_n(-x) d(-x) = (-1)^{m+n} \int_{-1}^{1} \frac{x^m}{\sqrt{1-x^2}} \mathcal{T}_n(x) dx = (-1)^{m+n} a_{m,n} \ ,$$

which implies that when $m+n$ is odd it satisfies $a_{m,n} = 0$. We next focus on the case when $m+n$ is even. We first consider two base cases:

- $n = 0, m = 2k$: we have $x^{2k} \geq 0$ for all $x \in [-1, 1]$ so $a_{m,n} = a_{2k,0} = \int_{-1}^{1} \frac{x^{2k}}{\sqrt{1-x^2}} dx \geq 0$.
- $n = 1, m = 2k+1$: we have $x^{2k+2} \geq 0$ for all $x \in [-1, 1]$ so $a_{m,n} = a_{2k+1,1} = \int_{-1}^{1} \frac{x^{2k+2}}{\sqrt{1-x^2}} dx \geq 0$.

As for general $n \geq 2$, we integrate by parts and have:

$$\begin{aligned}
a_{m,n} &= \int_{-\pi}^{0} \frac{\cos^m(\theta)}{\sin\theta} \cos(n\theta) d(\cos\theta) = \int_0^{\pi} \cos^m(\theta)\cos(n\theta) d\theta \\
&= \frac{1}{n} \cos^m(\theta) \sin(n\theta) \Big|_0^{\pi} - \frac{1}{n} \int_0^{\pi} (-m \sin(\theta) \cos^{m-1}(\theta)) \sin(n\theta) d\theta \\
&= \frac{m}{n} \int_0^{\pi} \sin(\theta) \cos^{m-1}(\theta) \sin(n\theta) d\theta \\
&= \frac{m}{n^2} \sin(\theta) \cos^{m-1}(\theta)(-\cos(n\theta)) \Big|_0^{\pi} - \frac{m}{n^2} \int_0^{\pi} (-(m-1)\cos^{m-2}\theta + m\cos^m\theta)(-\cos(n\theta)) d\theta \\
&= -\frac{m(m-1)}{n^2} a_{m-2,n} + \frac{m^2}{n^2} a_{m,n} \\
&\implies (m^2 - n^2) a_{m,n} = m(m-1) a_{m-2,n} \ . \qquad \text{(B.2)}
\end{aligned}$$

In particular, choosing $m = n$ in (B.2) we have $a_{n-2,n} = 0$, and this implies

$$\forall m \leq n\colon \quad a_{m-2,n} = \frac{m^2 - n^2}{m(m-1)} a_{m,n} = 0 \ .$$

As for $a_{n,n}$ for $n \geq 1$, we have

$$a_{n,n} = \int_{-1}^{1} \frac{x^n}{\sqrt{1-x^2}} \mathcal{T}_n(x) dx = \int_{-1}^{1} \frac{x^n}{\sqrt{1-x^2}} \frac{\mathcal{T}_{n+1}(x) + \mathcal{T}_{n-1}(x)}{2x} dx = \frac{1}{2}(a_{n-1,n+1} + a_{n-1,n-1}) = \frac{1}{2} a_{n-1,n-1}$$

and thus by induction we have $a_{n,n} \geq 0$. Using (B.2) again we conclude that

$$\forall m \geq n+2\colon \quad a_{m,n} = \frac{m(m-1)}{m^2 - n^2} a_{m-2,n} \geq 0 \ . \qquad \square$$

# C Proof of Lemma 5.4

We first note the following lemma which follows from Lemma 2.10 together with the aliasing Lemma 2.8:

**Lemma C.1.** *Suppose $f(z)$ is analytic on $\mathcal{E}_\rho$ and $|f(z)| \leq M$ on $\mathcal{E}_\rho$. Let $q_n(x) = \sum_{i=0}^{n} c_i \mathcal{T}_i(x)$ be the degree-$n$ Chebyshev interpolation of $f$, then*

$$\forall i \in \{0, 1, \ldots, n\}\colon \quad |c_i| \leq \frac{2M}{\rho + \sqrt{2\rho + \rho^2}} \left(1 + \rho + \sqrt{2\rho + \rho^2}\right)^{-i} \ .$$

Applying Lemma C.1 on $f(z) = \left(\frac{1+\kappa-z}{2}\right)^{-1/2}$, we have

**Lemma 5.4.** *Let $q_n(x) = \sum_{k=0}^{n} c_k \mathcal{T}_k(x)$ be the degree-$n$ Chebyshev interpolation of $f(x) = \left(\frac{1+\kappa-x}{2}\right)^{-1/2}$ on $[-1,1]$. Then,*

$$\forall i \in \{0,1,\dots,n\}: \quad |c_i| \le \frac{e\sqrt{32(i+1)}}{\kappa}\left(1+\kappa+\sqrt{2\kappa+\kappa^2}\right)^{-i}$$

*Proof of Lemma 5.4.* For each $i \in \{0,1,\dots,n\}$, consider a value $\rho \in [\kappa/2, \kappa)$ to be chosen later. We know that $f(z)$ is analytic and satisfies $|f(z)| \le \sqrt{\frac{2}{\kappa-\rho}}$ on $\mathcal{E}_\rho$. Using Lemma C.1 we have:

$$|c_i| \le \frac{\sqrt{\frac{8}{\kappa-\rho}}\left(1+\rho+\sqrt{2\rho+\rho^2}\right)^{-i}}{\rho+\sqrt{2\rho+\rho^2}} \le \frac{1}{\sqrt{\kappa}}\sqrt{\frac{8}{\kappa-\rho}}\left(1+\rho+\sqrt{2\rho+\rho^2}\right)^{-i} \quad , \tag{C.1}$$

where we used $\kappa \le 1$ in the second inequality. If we take $\rho = \kappa - \frac{\kappa}{4(i+1)}$, we have:

$$\begin{aligned}\left(\frac{1+\kappa+\sqrt{2\kappa+\kappa^2}}{1+\rho+\sqrt{2\rho+\rho^2}}\right)^i &= \left(1+\frac{\kappa-\rho+\sqrt{2\kappa+\kappa^2}-\sqrt{2\rho+\rho^2}}{1+\rho+\sqrt{2\rho+\rho^2}}\right)^i \\ &\le \left(1+(\kappa-\rho)\left(1+\frac{2+\kappa+\rho}{\sqrt{2\kappa}}\right)\right)^i \\ &\le \left(1+(\kappa-\rho)\frac{4}{\sqrt{\kappa}}\right)^i \le \left(1+\frac{1}{i+1}\right)^i \le e \ .\end{aligned}$$

Putting this back to (C.1), we have:

$$|c_i| \le \frac{\sqrt{32(i+1)}}{\kappa}\left(\frac{1+\kappa+\sqrt{2\kappa+\kappa^2}}{1+\rho+\sqrt{2\rho+\rho^2}}\right)^i \left(1+\kappa+\sqrt{2\kappa+\kappa^2}\right)^{-i} \le \frac{e\sqrt{32(i+1)}}{\kappa}\left(1+\kappa+\sqrt{2\kappa+\kappa^2}\right)^{-i} \ .$$

□

# D Appendix for Section 6

**Lemma 6.2.** $\vec{s}_N = \vec{b}_0 - \mathbf{M}\vec{b}_1$ *where*

$$\vec{b}_{N+1} \stackrel{\text{def}}{=} \vec{0}, \quad \vec{b}_N \stackrel{\text{def}}{=} \vec{c}_N, \quad \text{and} \quad \forall r \in \{N-1,\dots,0\}: \vec{b}_r \stackrel{\text{def}}{=} 2\mathbf{M}\vec{b}_{r+1} - \vec{b}_{r+2} + \vec{c}_r \in \mathbb{R}^d \ .$$

*Proof.* We write $\vec{s}_N = t^\top c$ where $t = (\mathcal{T}_0(\mathbf{M}),\dots,\mathcal{T}_N(\mathbf{M}))^\top$ and $c = (\vec{c}_0,\dots,\vec{c}_N)^\top$. Recall that the recursive formula of Chebyshev polynomial tells us

$$\mathbf{N}t \stackrel{\text{def}}{=} \begin{pmatrix} \mathbf{I} & & & & \\ -2\mathbf{M} & \mathbf{I} & & & \\ \mathbf{I} & -2\mathbf{M} & \mathbf{I} & & \\ & \ddots & \ddots & \ddots & \\ & & \mathbf{I} & -2\mathbf{M} & \mathbf{I} \end{pmatrix} \begin{pmatrix} \mathcal{T}_0(\mathbf{M}) \\ \mathcal{T}_1(\mathbf{M}) \\ \mathcal{T}_2(\mathbf{M}) \\ \vdots \\ \mathcal{T}_N(\mathbf{M}) \end{pmatrix} = \begin{pmatrix} \mathbf{I} \\ -\mathbf{M} \\ 0 \\ \vdots \\ 0 \end{pmatrix} \stackrel{\text{def}}{=} w \ .$$

In addition, it is easy to verify that the $\vec{b}_r$ sequence satisfies $\mathbf{N}^\top b = c$ if we denote by $b \stackrel{\text{def}}{=} (\vec{b}_0,\dots,\vec{b}_N)^\top$. Therefore, we have $\vec{s}_N = t^\top c = t^\top \mathbf{N}^\top b = w^\top b = \vec{b}_0 - \mathbf{M}\vec{b}_1$ as desired. □

**Fact D.1.** $\vec{b}_r = \sum_{k=r}^{N} \mathcal{U}_{k-r}(\mathbf{M})\vec{c}_k$ *for every* $r \in \{0,1,\dots,N+1\}$.

*Proof.* This can be deduced directly from the recursive formula of Chebyshev polynomials of the second kind. See for instance Equation (3.120) of [14]. □

## D.1 Proof of Theorem 6.4

**Theorem 6.4.** *For every $N \in \mathbb{N}^*$, suppose the eigenvalues of $\mathbf{M}$ are in $[a,b]$ and suppose there are parameters $C_U \geq 1, C_T \geq 1, \rho \geq 1, C_c \geq 0$ satisfying*

$$\forall k \in \{0,1,\dots,N\}\colon \Big\{\rho^k \|\vec{c}_k\| \leq C_c \quad \bigwedge \quad \forall x \in [a,b]\colon \quad |\mathcal{T}_k(x)| \leq C_T \rho^k \text{ and } |\mathcal{U}_k(x)| \leq C_U \rho^k \Big\} \ .$$

*Then, if the inexact backward recurrence in Def. 6.3 is applied with $\varepsilon \leq \frac{1}{4NC_U}$, we have*

$$\|\widehat{s}_N - \vec{s}_N\| \leq \varepsilon \cdot 2(1+2NC_T)NC_UC_c \ .$$

*Proof of Theorem 6.4.* We first note that according to $\vec{b}_n = \sum_{k=n}^N \mathcal{U}_{k-n}(\mathbf{M})\vec{c}_k$ from Fact D.1, we have

$$\forall n \in \{0,1,\dots,N\}\colon \quad \|\vec{b}_n\| \leq \sum_{k=n}^{N} \big\|\mathcal{U}_{k-n}(\mathbf{M})\big\|_2 \|\vec{c}_k\| \leq (N-n+1)\cdot \rho^{-n}\cdot C_UC_c \ . \tag{D.1}$$

Denoting by $\vec{\eta}_r \stackrel{\text{def}}{=} \mathcal{M}\big(\widehat{b}_r\big) - \mathbf{M}\widehat{b}_r$, we have

$$\forall r \in \{N-1,\dots,0\}\colon \widehat{b}_r = 2\mathbf{M}\widehat{b}_{r+1} - \widehat{b}_{r+2} + \vec{c}_r + 2\vec{\eta}_{r+1} \quad \text{and} \quad \widehat{s}_N = \widehat{b}_0 + \mathbf{M}\widehat{b}_1 + \vec{\eta}_1 \ ,$$

and therefore if we denote by $(\delta\vec{b})_r = \widehat{b}_r - \vec{b}_r$, we have

$$(\delta\vec{b})_{N+1} = 0, \quad (\delta\vec{b})_N = 0, \quad \text{and} \quad \forall r \in \{N-1,\dots,0\}\colon (\delta\vec{b})_r = 2\mathbf{M}(\delta\vec{b})_{r+1} - (\delta\vec{b})_{r+2} + 2\vec{\eta}_{r+1} \ .$$

In other words, the $\{(\delta\vec{b})_r\}_r$ sequence also satisfies the recursive formula in Lemma 6.2 where $\vec{c}_k$ is replaced with $2\vec{\eta}_{k+1}$. This implies, according to Lemma 6.2,

$$(\delta\vec{b})_0 - \mathbf{M}(\delta\vec{b})_1 = 2\sum_{k=0}^{N-1} \mathcal{T}_k(\mathbf{M})\vec{\eta}_{k+1}$$

and therefore

$$\widehat{s}_N - \vec{s}_N = (\delta\vec{b})_0 - \mathbf{M}(\delta\vec{b})_1 + \vec{\eta}_1 = \vec{\eta}_1 + 2\sum_{k=0}^{N-1} \mathcal{T}_k(\mathbf{M})\vec{\eta}_{k+1} \tag{D.2}$$

At the same time, applying Fact D.1 on sequence $\{(\delta\vec{b})_r\}_r$, we have

$$(\delta\vec{b})_r = 2\sum_{k=r}^{N-1} \mathcal{U}_{k-r}(\mathbf{M})\vec{\eta}_{k+1} \tag{D.3}$$

Now we are ready to prove that, as long as $\varepsilon \leq \frac{1}{4NC_U}$, it satisfies

$$\forall k \in [N]\colon \quad \|\vec{\eta}_k\| \leq \varepsilon \cdot \rho^{-k}(2NC_UC_c) \quad \text{and} \quad \|(\delta\vec{b})_{k-1}\| \leq \varepsilon\cdot\rho^{-k}(4N^2C_U^2C_c)$$

We prove this by reverse double induction.

- In the base case, $\|\vec{\eta}_N\| \leq \varepsilon\|\widehat{b}_N\| = \varepsilon\|\vec{b}_N\| \leq \varepsilon\rho^{-N}C_UC_c$ where the first inequality uses our assumption on $\mathcal{M}$ and the second uses (D.1).
- Suppose the upper bound $\|\vec{\eta}_k\| \leq \varepsilon\cdot\rho^{-k}(2NC_UC_c)$ holds for every $k \geq k_0$, then

$$\begin{aligned}\|(\delta\vec{b})_{k_0-1}\| &\leq 2\sum_{k=k_0-1}^{N-1} \Big\|\mathcal{U}_{k-k_0+1}(\mathbf{M})\vec{\eta}_{k+1}\Big\| \leq 2C_U\rho^{-k_0}\cdot \sum_{k=k_0-1}^{N-1} \|\rho^{k+1}\vec{\eta}_{k+1}\| \\ &\leq 2NC_U\rho^{-k_0}\cdot(\varepsilon\cdot 2NC_UC_c) = \varepsilon\cdot\rho^{-k_0}(4N^2C_U^2C_c) \ .\end{aligned}$$

  Above, the first inequality is by (D.3) and triangle inequality, the second is by the definition of $C_U$, the third is by inductive assumption.

- Suppose the upper bound $\|(\delta\vec{b})_{k-1}\| \leq \varepsilon \cdot \rho^{-k}(4N^2C_U^2C_c)$ holds for $k = k_0 + 1$, then

$$\begin{aligned}\|\vec{\eta}_{k_0}\| \leq \varepsilon\|\widehat{b}_{k_0}\| \leq \varepsilon\big(\|\vec{b}_{k_0}\| + \|(\delta\vec{b})_{k_0}\|\big) &\leq \varepsilon\rho^{-k_0}\big(NC_UC_c + 4\varepsilon\rho^{-1}N^2C_U^2C_c\big)\\ &= \varepsilon\rho^{-k_0}NC_UC_c(1 + 4\rho^{-1}\varepsilon NC_U) \leq 2\rho^{-k_0}NC_UC_c\ .\end{aligned}$$

  Above, the first inequality is by our assumption on $\mathcal{M}$, the second is by triangle inequality, the third is by (D.1) and our inductive assumption, and the last is by our assumption on $\varepsilon$.

Finally, using (D.2), we have

$$\begin{aligned}\|\widehat{s}_N - \vec{s}_N\| \leq \|\vec{\eta}_1\| + 2\sum_{k=0}^{N-1}\big\|\mathcal{T}_k(\mathbf{M})\|_2\|\vec{\eta}_{k+1}\| &\leq \varepsilon(2NC_UC_c) + 2N\varepsilon C_T(2NC_UC_c)\\ &\leq \varepsilon \cdot 2(1 + 2NC_T)NC_UC_c\ .\end{aligned}$$

□

# E Appendix for Section 7

**Fact 7.1.** $(\mathbf{P}_\lambda)\chi = \frac{\mathbf{I}+\mathrm{sgn}(\mathbf{S})}{2}$ *where* $\mathbf{S} \stackrel{\mathrm{def}}{=} 2(\mathbf{A}^\top\mathbf{A} + \lambda\mathbf{I})^{-1}\mathbf{A}^\top\mathbf{A} - \mathbf{I} = (\mathbf{A}^\top\mathbf{A} + \lambda\mathbf{I})^{-1}(\mathbf{A}^\top\mathbf{A} - \lambda\mathbf{I})$.

*Proof.* This is so because $\mathbf{S}$ shares the same eigenspace as $\mathbf{A}^\top\mathbf{A}$ and maps all the eigenvalues of $\mathbf{A}^\top\mathbf{A}$ above threshold $\lambda$ to eigenvalues of $\mathbf{S}$ between 0 and 1, and all the eigenvalues below $\lambda$ to eigenvalues of $\mathbf{S}$ between $-1$ and 0. Therefore, if applied to function $\frac{\mathrm{sgn}(x)+1}{2}$, we have that $\frac{\mathrm{sgn}(\mathbf{S})+\mathbf{I}}{2}$ zeros out all the eigenvalues of $\mathbf{S}$ between $-1$ and 0, and thus equivalently zeros out all the eigenvalues of $\mathbf{A}^\top\mathbf{A}$ below threshold $\lambda$. This is exactly the same as the projection matrix $\mathbf{P}_\lambda$. □

## E.1 Proof of Theorem 7.3

**Theorem 7.3** (restated)**.** *Given* $\mathbf{A} \in \mathbb{R}^{d'\times d}$ *and* $\lambda, \gamma \in (0, 1)$*, assume that the singular values of* $\mathbf{A}$ *are in the range* $[0, \sqrt{(1-\gamma)\lambda}] \cup [\sqrt{(1+\gamma)\lambda}, 1]$*. Given* $\chi \in \mathbb{R}^d$ *and* $b \in \mathbb{R}^{d'}$*, denote by* $\xi^* = \mathbf{P}_\lambda\chi$ *and* $x^* = (\mathbf{A}^\top\mathbf{A})^{-1}\mathbf{P}_\lambda\mathbf{A}^\top b$ *the exact PCP and PCR solutions. If* `ApxRidge` *is an* $\varepsilon'$*-approximate ridge regression solver, then*

$$\textit{the output}\quad \xi \leftarrow \mathtt{QuickPCP}(\mathbf{A}, \chi, \lambda, \gamma, n) \quad \textit{satisfies}\quad \|\xi^* - \xi\| \leq \varepsilon\|\chi\|$$

*if* $n = \Theta\big(\gamma^{-1}\log\frac{1}{\gamma\varepsilon}\big)$ *and* $\log(1/\varepsilon') = \Theta\big(\log\frac{1}{\gamma\varepsilon}\big)$*;*

$$\textit{the output}\quad x \leftarrow \mathtt{QuickPCR}(\mathbf{A}, b, \lambda, \gamma, n, m) \quad \textit{satisfies}\quad \|x - x^*\| \leq \varepsilon\|b\|$$

*if* $n = \Theta\big(\gamma^{-1}\log\frac{1}{\gamma\lambda\varepsilon}\big)$*,* $m = \Theta\big(\log\frac{1}{\gamma\varepsilon}\big)$ *and* $\log(1/\varepsilon') = \Theta\big(\log\frac{1}{\gamma\lambda\varepsilon}\big)$*.*

*Proof of Theorem 7.3.* The eigenvalues of $\mathbf{S} \stackrel{\mathrm{def}}{=} (\mathbf{A}^\top\mathbf{A} + \lambda\mathbf{I})^{-1}(\mathbf{A}^\top\mathbf{A} - \lambda\mathbf{I})$ are in the range

$$\Big[-1, -\frac{(1+\gamma)-1}{1+(1-\gamma)}\Big] \cup \Big[\frac{1-(1-\gamma)}{1+(1+\gamma)}, \frac{1-\lambda}{1+\lambda}\Big] \subseteq \big[-1, -\alpha\big] \cup \big[\alpha, 1\big]\ .$$

because $\alpha = \gamma/(2+\gamma)$. Therefore, according to Theorem 5.1, $g_n(\mathbf{S})\chi$ satisfies $\|g_n(\mathbf{S})\chi - \mathrm{sgn}(\mathbf{S})\chi\| \leq \varepsilon\|\chi\|$ for every $\chi \in \mathbb{R}^d$ which in turns implies $\|\frac{1}{2}(g_n(\mathbf{S}) + \mathbf{I})\chi - \mathbf{P}_\lambda\chi\| \leq \varepsilon\|\chi\|$.

We now analyze stability. Denote by $\mathbf{M} = (1+\kappa)\mathbf{I} - 2\mathbf{S}^2$ and recall that $g_n(\mathbf{S}) = \mathbf{S}q_n(\mathbf{M}) = \mathbf{S}q_n\big((1+\kappa)\mathbf{I} - 2\mathbf{S}^2\big)$ where $\kappa = 2\alpha^2$. We wish to apply Theorem 6.4 to show that $q_n(\mathbf{M})\chi$ can be computed in a stable manner and therefore $g_n(\mathbf{S})\chi$ as well. We verify the assumptions of Theorem 6.4 below:

- Since ApxRidge is $\varepsilon'$-approximate (see Def. 2.3), we have that Line 6 of QuickPCP corresponds to an approximate algorithm

$$\mathcal{M}(\chi) = (1+\kappa)\chi - 2\texttt{MultS}(\mathbf{A}, \lambda, \texttt{MultS}(\mathbf{A}, \lambda, \chi))$$

satisfying $\|\mathbf{M}\chi - \mathcal{M}(\chi)\| \le O(\varepsilon')\|\chi\|$ for every vector $\chi$.

- Recall that $q_n(\cdot)$ is a Chebyshev sum with coefficients at most $O(1/\sqrt{\kappa}) = O(1/\alpha) = O(1/\gamma)$ according to Def. 2.7. Thus, we can choose $\rho = 1$ and $C_c = O(1/\gamma)$ in Theorem 6.4.
- Since the eigenvalues of $\mathbf{M}$ are in $[-1, 1]$ and $|\mathcal{T}_k(x)| \le 1$ and $|\mathcal{U}_k(x)| \le n+1$ for every $x \in [-1, 1]$ (see Fact 2.6), we can choose $C_T = 1$ and $C_U = n+1$ in Theorem 6.4.

The conclusion of Theorem 6.4 tells us that our approximate backward recurrence in QuickPCP computes $g_n(\mathbf{S})\chi$ up to an accuracy $O(\varepsilon'\gamma^{-1}n^3) \cdot \|\chi\|$. In other words, as long as $\log(1/\varepsilon') \le O(\log \frac{n}{\varepsilon\gamma})$, we can approximately compute $\frac{1}{2}(g_n(\mathbf{S}) + \mathbf{I})\chi$ within accuracy $O(\varepsilon) \cdot \|\chi\|$.

Combining everything above, we conclude that choosing $n = \Theta(\gamma^{-1}\log(1/\gamma\varepsilon))$ and $\log(1/\varepsilon') = \Theta(\log \frac{n}{\varepsilon\gamma}) = \Theta(\log \frac{1}{\varepsilon\gamma})$, we can satisfy $\|\xi^* - \xi\| \le \varepsilon\|\chi\|$.

As for the PCR guarantee, we simply replace $\varepsilon$ with $\varepsilon \cdot \sqrt{\lambda}/m^2$ and $\chi$ with $\mathbf{A}^\top b$ in the above analysis. Then we apply Lemma 3.4, and conclude that choosing $n = \Theta(\gamma^{-1}\log(1/\gamma\lambda\varepsilon))$, $m = \Theta(\log(1/\varepsilon\gamma))$, and $\log(1/\varepsilon') = \Theta(\log \frac{1}{\varepsilon\gamma})$, it satisfies $\|x^* - x\| \le \varepsilon\|b\|$. □

## E.2 Proof of Theorem 7.4

**Theorem 7.4** (restated). *Given $\mathbf{A} \in \mathbb{R}^{d' \times d}$, $\lambda \in (0,1)$, and $\gamma \in (0, 2/3]$, assume that the singular values of $\mathbf{A}$ are no more than 1. Given $\chi \in \mathbb{R}^d$ and $b \in \mathbb{R}^{d'}$, and suppose* ApxRidge *is an $\varepsilon'$-approximate ridge regression solver, then*

*the output* $\xi \leftarrow \texttt{QuickPCP}(\mathbf{A}, \chi, \lambda, \gamma, n)$ *is $(\gamma, \varepsilon)$-approximate PCP*

*if $n = \Theta\big(\gamma^{-1}\log\frac{1}{\gamma\varepsilon}\big)$ and $\log(1/\varepsilon') = \Theta\big(\log\frac{1}{\gamma\varepsilon}\big)$;*

*the output* $x \leftarrow \texttt{QuickPCR}(\mathbf{A}, b, \lambda, \gamma, n, m)$ *is $(\gamma, \varepsilon)$-approximate PCR*

*if $n = \Theta\big(\gamma^{-1}\log\frac{1}{\gamma\lambda\varepsilon}\big)$, $m = \Theta\big(\log\frac{1}{\gamma\varepsilon}\big)$ and $\log(1/\varepsilon') = \Theta\big(\log\frac{1}{\gamma\lambda\varepsilon}\big)$.*

*Proof of Theorem 7.4.* Consider the same $\mathbf{S} = (\mathbf{A}^\top\mathbf{A} + \lambda\mathbf{I})^{-1}(\mathbf{A}^\top\mathbf{A} - \lambda\mathbf{I})$, $\alpha = \gamma/(2+\gamma)$, and $\kappa = 2\alpha^2$ as before. We observe that $\mathbf{A}^\top\mathbf{A}$ and $\mathbf{S}$ share the same eigenspace. Furthermore, the eigenvalues of $\mathbf{A}^\top\mathbf{A}$ in the range

$$(1)\colon [(1+\gamma)\lambda, 1] \qquad (2)\colon [0, (1-\gamma)\lambda] \qquad (3)\colon \big((1-\gamma)\lambda, (1+\gamma)\lambda\big) \tag{E.1}$$

respectively map to the eigenvalues of $\mathbf{S}$ in the range[16]

$$(1)\colon [\alpha, 1] \qquad (2)\colon [-1, -\alpha] \qquad (3)\colon (-1, 1)$$

Let us now write $\chi = \sum_{i=1}^3 \sum_{k \in \Lambda_i} \beta_k \nu_k$ where $\Lambda_i \subseteq [d]$ consists of the indices $k$ where $\lambda_k$ is in the $i$-th interval in (E.1), and $\beta_k \in \mathbb{R}$ is the weight. We thus have

$$\xi' \stackrel{\text{def}}{=} \frac{g_n(\mathbf{S}) + \mathbf{I}}{2}\chi = \sum_{i=1}^3 \sum_{k \in \Lambda_i} \frac{g_n(\lambda_k) + \lambda_k}{2}\beta_k \nu_k \ .$$

[16]More precisely, given transformation $f\colon x \mapsto \frac{x-\lambda}{x+\lambda}$, it satisfies (1) $f\big([(1+\gamma)\lambda, 1]\big) \subseteq [\alpha, 1]$, (2) $f\big([0, (1-\gamma)\lambda]\big) \subseteq [-1, -\alpha]$, and (3) $f\big(((1-\gamma)\lambda, (1+\gamma)\lambda)\big) \subseteq (-1, 1)$.

Since for every $k \in \Lambda_1 \cup \Lambda_2$ it satisfies $\lambda_k \in [-1, -\alpha] \cup [\alpha, 1]$, we can apply Theorem 5.1 (and using $n \geq \frac{1}{\sqrt{2}\alpha} \log \frac{3}{\varepsilon\alpha^2}$):

1. $\|\mathbf{P}_{(1+\gamma)\lambda}(\xi' - \chi)\| = \| \sum_{k\in\Lambda_1} \big(\frac{g_n(\lambda_k)+\lambda_k}{2} - 1\big)\beta_k \nu_k\| \leq \varepsilon\|\chi\|$.
2. $\|(\mathbf{I} - \mathbf{P}_{(1-\gamma)\lambda})\xi'\| = \| \sum_{k\in\Lambda_2} \big(\frac{g_n(\lambda_k)+\lambda_k}{2}\big)\beta_k \nu_k\| \leq \varepsilon\|\chi\|$.
3. $\forall k \in \Lambda_3$, $|\langle \nu_i, \xi' - \chi\rangle| = \big|\frac{g_n(\lambda_k)+\lambda_k}{2} - 1\big| \cdot |\beta_k| \leq |\beta_k| = |\langle \nu_i, \chi\rangle|$. (Here, the last inequality is because if $\lambda_k \geq 0$ then $g_n(\lambda_k) + \lambda_k \in [\lambda_k, 1 + \lambda_k]$ and if $\lambda_k < 0$ then $g_n(\lambda_k) + \lambda_k \in [\lambda_k - 1, \lambda_k]$.)

Note that these two guarantees correspond to the three properties for approximate PCP (see Def. 3.1), and thus we are left to deal with stability by applying Theorem 6.4. In other words, denoting by $\mathbf{M} = (1+\kappa)\mathbf{I} - 2\mathbf{S}^2$ and recalling that $g_n(\mathbf{S}) = \mathbf{S}q_n(\mathbf{M})$, we wish to apply Theorem 6.4 to show that $q_n(\mathbf{M})\chi$ can be computed in a stable manner and therefore $\frac{q_n(\mathbf{S})+\mathbf{I}}{2}\chi$ as well. We verify the assumptions of Theorem 6.4 below:

- As before, Line 6 of `QuickPCP` corresponds to an approximate algorithm $\mathcal{M}(\chi)$ satisfying $\|\mathbf{M}\chi - \mathcal{M}(\chi)\| \leq O(\varepsilon')\|\chi\|$ for every vector $\chi$.
- $q_n(\cdot)$ is a Chebyshev sum with coefficients satisfying $|c_i| \leq O(\sqrt{i}/\kappa)\big(1 + \kappa + \sqrt{2\kappa + \kappa^2}\big)^{-i}$ according to Lemma 5.4. Therefore, we can choose $\rho = 1 + \kappa + \sqrt{2\kappa + \kappa^2}$ and $C_c = O(n/\kappa) = O(n/\gamma^2)$ in Theorem 6.4.
- Since the eigenvalues of $\mathbf{M}$ are in $[-1, 1 + \kappa]$, we have for every $x \in [-1, 1 + \kappa]$, it satisfies $|\mathcal{T}_k(x)| \leq (1 + \kappa + \sqrt{2\kappa + \kappa^2})^k$ and $|\mathcal{U}_k(x)| \leq \frac{1}{2\sqrt{2\kappa+\kappa^2}}(1 + \kappa + \sqrt{2\kappa + \kappa^2})^{n+1}$. Therefore, we can choose $C_T = 1$ and $C_U = O(\frac{1}{\kappa})$ in Theorem 6.4.

Finally, the conclusion of Theorem 6.4 tells us that our approximate backward recurrence in `QuickPCP` computes $q_n(\mathbf{S})\chi$ up to an accuracy $O(\varepsilon'\gamma^{-4}n^3) \cdot \|\chi\|$. In other words, as long as $\log(1/\varepsilon') \leq O(\log \frac{n}{\varepsilon\gamma})$, we can approximately compute $\xi' = \frac{1}{2}(g_n(\mathbf{S}) + \mathbf{I})\chi$ within accuracy $\varepsilon \cdot \|\chi\|$, or equivalently $\|\xi - \xi'\| \leq \varepsilon\|\chi\|$. Together with our analysis at the beginning of the proof, we have

1. $\|\mathbf{P}_{(1+\gamma)\lambda}(\xi - \chi)\| \leq 2\varepsilon\|\chi\|$.
2. $\|(\mathbf{I} - \mathbf{P}_{(1-\gamma)\lambda})\xi\| \leq 2\varepsilon\|\chi\|$.
3. $\forall k \in \Lambda_3$, $|\langle \nu_i, \xi - \chi\rangle| \leq |\beta_k| = |\langle \nu_i, \chi\rangle| + \varepsilon\|\chi\|$.

This finishes proving that $\xi$ is an $(\gamma, O(\varepsilon))$-approximate PCP solution when $n = \Theta\big(\gamma^{-1} \log \frac{1}{\gamma\varepsilon}\big)$ and $\log(1/\varepsilon') = \Theta\big(\log \frac{1}{\gamma\varepsilon}\big)$.

As for the PCR guarantee, we simply replace $\varepsilon$ with $\varepsilon \cdot \lambda/m^2$ and $\chi$ with $\mathbf{A}^\top b$ in the above analysis. Then we apply Lemma 3.5, and conclude that choosing $n = \Theta(\gamma^{-1} \log(1/\varepsilon\lambda\gamma))$, $m = \Theta(\log(1/\varepsilon\gamma))$, and $\log(1/\varepsilon') = \Theta(\log \frac{1}{\varepsilon\gamma})$, it satisfies that $x$ is a $(\gamma, \varepsilon)$-approximate PCR solution. □